\documentclass{lmcs} %%% last changed 2014-08-20


\usepackage{hyperref}
\usepackage{subfigure}
\usepackage{booktabs}
\usepackage{cite}
\usepackage{graphicx}

\theoremstyle{plain} %\crefname{satz}{Satz}{S\"atze}
\begin{document}
		
		\title[YOLO algorithm with HAFPN for solder joint defect detection]{YOLO algorithm with hybrid attention feature pyramid network for solder joint defect detection %\texorpdfstring{\MakeLowercase{\texttt{lmcs.cls}}}{lmcs.cls}\rsuper*\\Version of
	%		2024-01-02
			}
	%	\titlecomment{{\lsuper*}OPTIONAL comment concerning the title, \eg,
	%		if a variant or an extended abstract of the paper has appeared elsewhere.}
		\thanks{*Corresponding author}	%optional
		
		% affiliations are numbered automatically with a, b, c (see below)
		% use the optional argument to indicate the affiliation(s) of each author
		% omit the argument if there is only one author, or only one affiliation
		\author[L.~Ang]{Li Ang\lmcsorcid{0009-0008-2964-288X}}[a,b]
		\author[SKNA.~Rahim]{Siti Khatijah Nor Abdu Rahim\lmcsorcid{0000-0002-1212-8867}}[a]
		\author[R.~Hamzah]{Raseeda Hamzah$^{\ast}$\lmcsorcid{0000-0002-9202-125X}}[c]
		\author[R.~Aminuddin]{Raihah Aminuddin\lmcsorcid{0000-0003-3058-7458}}[c]
		\author[G.~Yousheng]{Gao Yousheng\lmcsorcid{0009-0005-9519-3240}}[a,b]
		
		% affiliation 1 (automatically numbered a)
		\address{College of Computing, Informatics and Mathematics, Universiti Teknologi MARA (UiTM), Shah Alam, Selangor, Malaysia}	%optional
		% write emails for all authors having that affiliation
		\email{2022667284@student.uitm.edu.my, sitik781@uitm.edu.my}  %optional
		
		% affiliation 2 (automatically numbered b)
		\address{College of Information Engineering, Jiujiang Vocational University, Jiu Jiang, Jiang Xi, China}	%optional
		\email{2022667284@student.uitm.edu.my}  %optional
		
		% affiliation 3 (automatically numbered b)
		\address{College of Computing, Informatics and Mathematics, Universiti Teknologi MARA (UiTM) Melaka Branch}	%optional
		\email{raseeda@uitm.edu.my, raihah1@uitm.edu.my}  %optional
		
		%% etc.
		
		%% required for running head on odd and even pages, use suitable
		%% abbreviations in case of long titles and many authors:
		
		%%%%%%%%%%%%%%%%%%%%%%%%%%%%%%%%%%%%%%%%%%%%%%%%%%%%%%%%%%%%%%%%%%%%%%%%%%%
		
		%% the abstract has to PRECEDE the command \maketitle:
		%% be sure not to issue the \maketitle command twice!
		
		\begin{abstract}
			Traditional manual detection for solder joint defect is no longer applied during industrial production due to low efficiency, inconsistent evaluation, high cost and lack of real-time data. A new approach has been proposed to address the issues of low accuracy, high false detection rates and computational cost of solder joint defect detection in surface mount technology of industrial scenarios. The proposed solution is a hybrid attention mechanism designed specifically for the solder joint defect detection algorithm to improve quality control in the manufacturing process by increasing the accuracy while reducing the computational cost. The hybrid attention mechanism comprises a proposed enhanced multi-head self-attention and coordinate attention mechanisms increase the ability of attention networks to perceive contextual information and enhances the utilization range of network features. The coordinate attention mechanism enhances the connection between different channels and reduces location information loss. The hybrid attention mechanism enhances the capability of the network to perceive long-distance position information and learn local features. The improved algorithm model has good detection ability for solder joint defect detection, with mAP reaching 91.5\%, 4.3\% higher than the You Only Look Once version 5 algorithm and better than other comparative algorithms. Compared to other versions, mean Average Precision, Precision, Recall, and Frame per Seconds indicators have also improved. The improvement of detection accuracy can be achieved while meeting real-time detection requirements.
		\end{abstract}
		
		\maketitle
		
		%% start the paper here:
		\section{Introduction}\label{S:one}
		
		Surface mount device (SMD) pins are prone to lead welding defects in automatic production, such as insufficient defect, and foot shifting defect, as in Figure 1. In solder joint defect detection, traditional manual detection is no longer adapted to the development of industrial production. Manual detection is low efficiency, inconsistent evaluation, high cost and lack of real-time data.		
		
		Computer vision is a combination of computer hardware and software working together with industrial cameras and source of lights for capturing an image. It is utilized in a variety of industrial scenes to automate manufacturing and improve product quality. The solder joint defect detection system based on computer vision has the characteristics of real-time, continuous, and contact-less. This approach can take the place of manual detection and enhance the accuracy of results. At present, computer vision has been frequently employed in defect detection. Therefore, the use of computer vision for detecting solder joint defects has become a mainstream trend. In recent years, deep learning technology, that is one stream of computer vision has been developed rapidly. The research being developed for automated solder joint defect detection is still lacking. The methods for solder joint defect detection can be separated into three main groups, feature-based methods \cite{01,02}, statistical methods \cite{03,04}, and deep learning methods \cite{05,06,07}. The deep learning method can learn effective information and rules from the solder joint image due to its CNN structure \cite{08}. It can solve the problem that defective features are difficult to extract by artificially designed rules. The structures of deep learning neural network (DNN) can be seen in two folds: one is single-stage and the other is two-stage networks. Although the two-stage DNN is more accurate compared to the first stage, the shallow feature needs to be carefully utilized to avoid missing information during the feature extraction stage that will lead to lower detection rate. Its real-time performance is also poor, and it is not suitable for an industrial production environment. On the other hand, the one-stage method represented has good real-time performance and fast detection speed, but the detection effect is not good for small defect area and low-resolution images \cite{09}. In the feature extraction module of defect detection, the target feature information is lost too much, and the detection rate of small defect is not ideal, leading to serious issues of missed detection \cite{10}. The deep learning method uses deep neural network to extract features, but as the network layers of the deep neural network deepen, some shallow information is easily lost, leading to missed detection of small-sized object. To address this issue, the multi-scale feature fusion method is adopted to fuse deep and shallow features in the feature extraction process, enhancing information transmission between different network layers. Therefore, optimizing feature fusion methods can promote the improvement of detection accuracy for small-sized object.
		
		\begin{figure}[]
			\centering
			\subfigure[]{
				\includegraphics[scale=0.15]{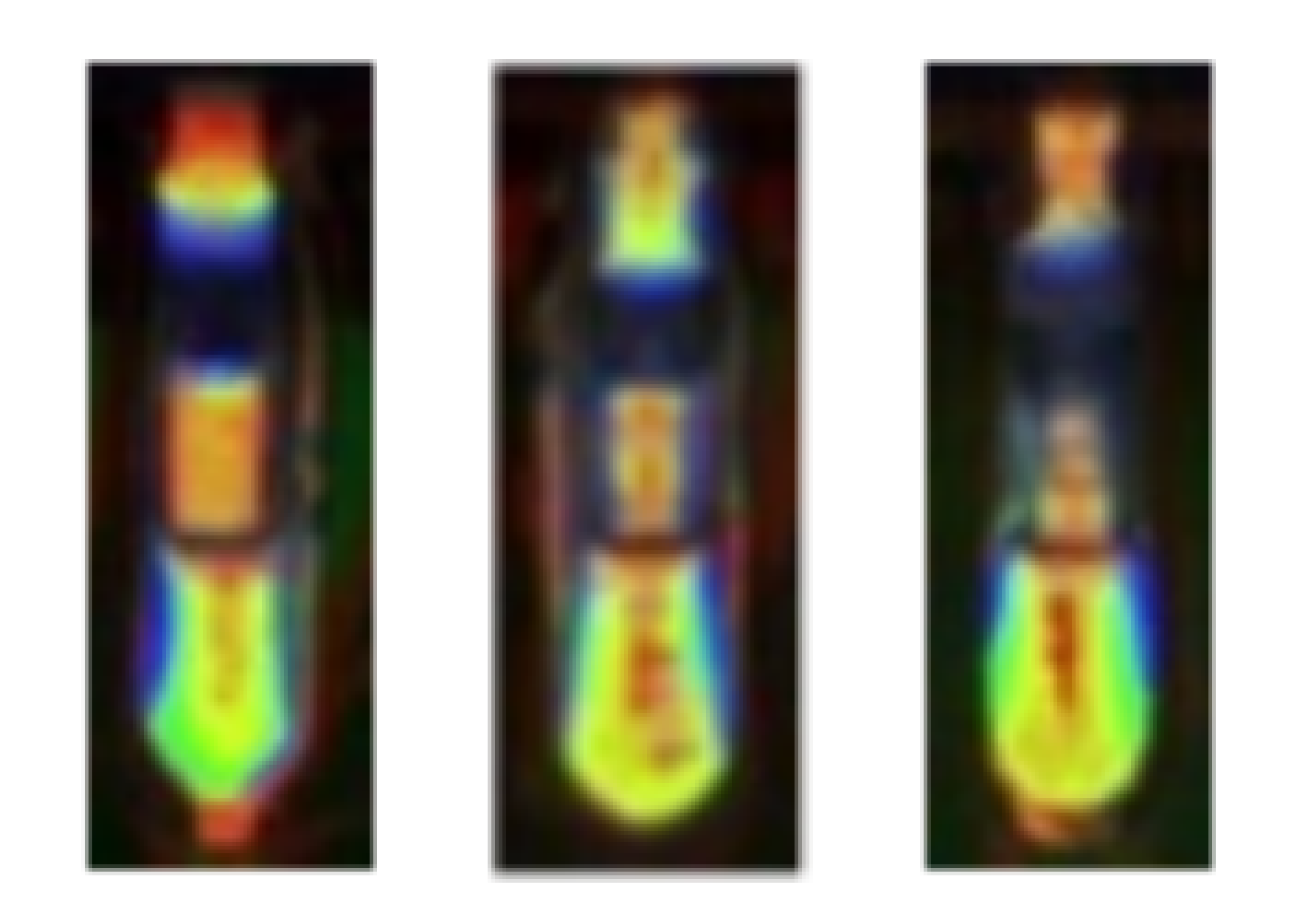}}
			\subfigure[]{
				\includegraphics[scale=0.15]{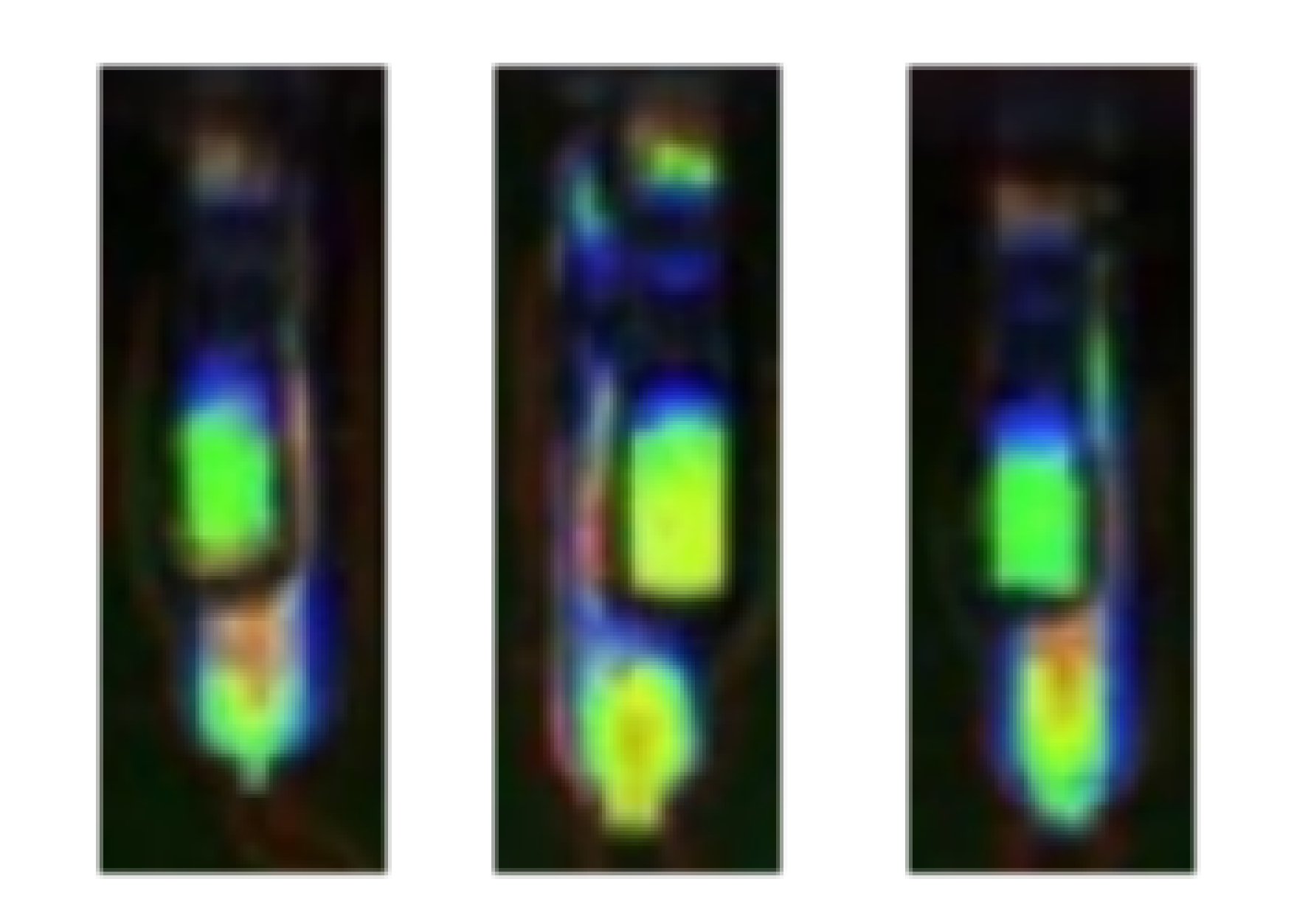}}
			\subfigure[]{
				\includegraphics[scale=0.15]{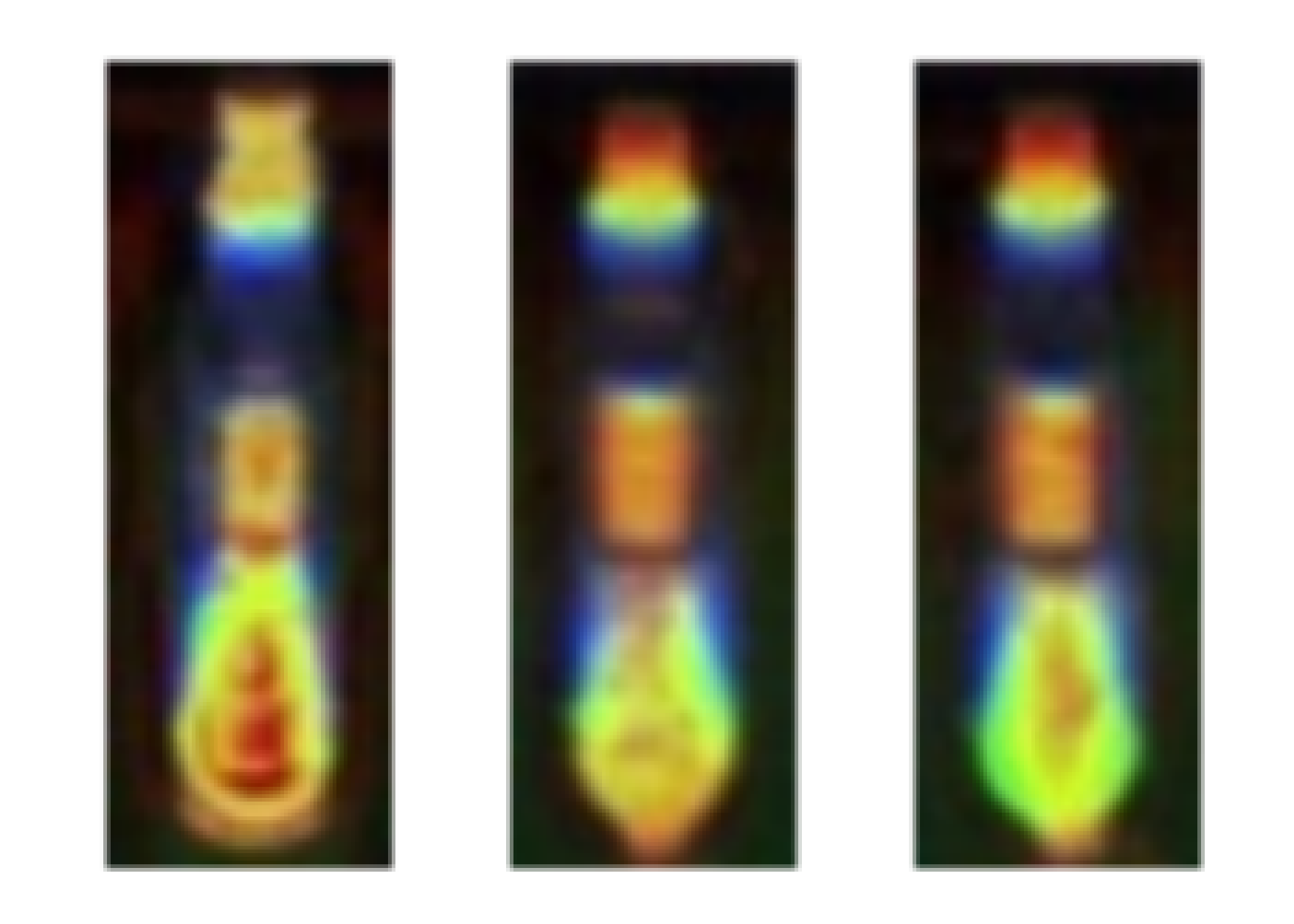}}
			\caption{(a) Insufficient, (b) Foot shifting, and (c) Qualified solder joint samples in dataset}
			\label{fig_1}
		\end{figure}
		
		Feature Pyramid Network (FPN) \cite{11} obtains feature images of different scales through multiple up-sampling of the input image. It integrates the abstract semantic information extracted from the high-level with the specific details, like the low-level contour texture in the feature extraction process from top to bottom to fulfill the goal of feature extraction enhancement. However, although FPN has systematically extracted the low-level and high-level features, its feature fusion capability still cannot meet the requirements, making it difficult to retain shallow features information.
		
		To address the missing information between high-level and low-level features, Liu et al. \cite{12} designed Path Aggregation Network (PANet) to connect a bottom-up enhancement path at the bottom of the feature pyramid. This process is done to shorten the transmission path of information fusion, to feed the fusion network location data with fine-grained features to increase the feature pyramid architecture's detection capacity. Bidirectional Feature Pyramid Network (BiFPN) \cite{13} is founded on the PANet which the node with only one input is removed to reduce the amount of parameter calculation. Directly connect the input and output layers of features through an additional skip transmission path to enhance the fusion ability of shallow features. BiFPN assigns weights to each layer of adaptive learning, and through the allocation of weights, the network perceives the importance of different levels. Multi scale feature fusion is widely used in small object detection. It significantly improves the detection performance of small objects by combining high-level semantic information with low-level detailed information. However, the construction of the FPN is mainly divided into cross-layer connection and parallel branch. Although the mechanism increases the performance, it adds additional parameter calculation and storage space. Therefore, the investigation of designing a pyramid feature network architecture that is able to enhance the feature fusion capability of the defect detector is required. We propose a hybrid attention mechanism to improve the feature fusion ability of feature pyramid networks. We applied the enhanced FPN to the YOLOv5 detection model. This paper designs comparative experiments and ablation experiments to verify the effectiveness of the proposed method on the solder joint defect dataset. The overall process flowchart of the paper is shown in Figure 2. 
		
		\begin{figure}
			\centering
			\includegraphics[width=0.5\linewidth]{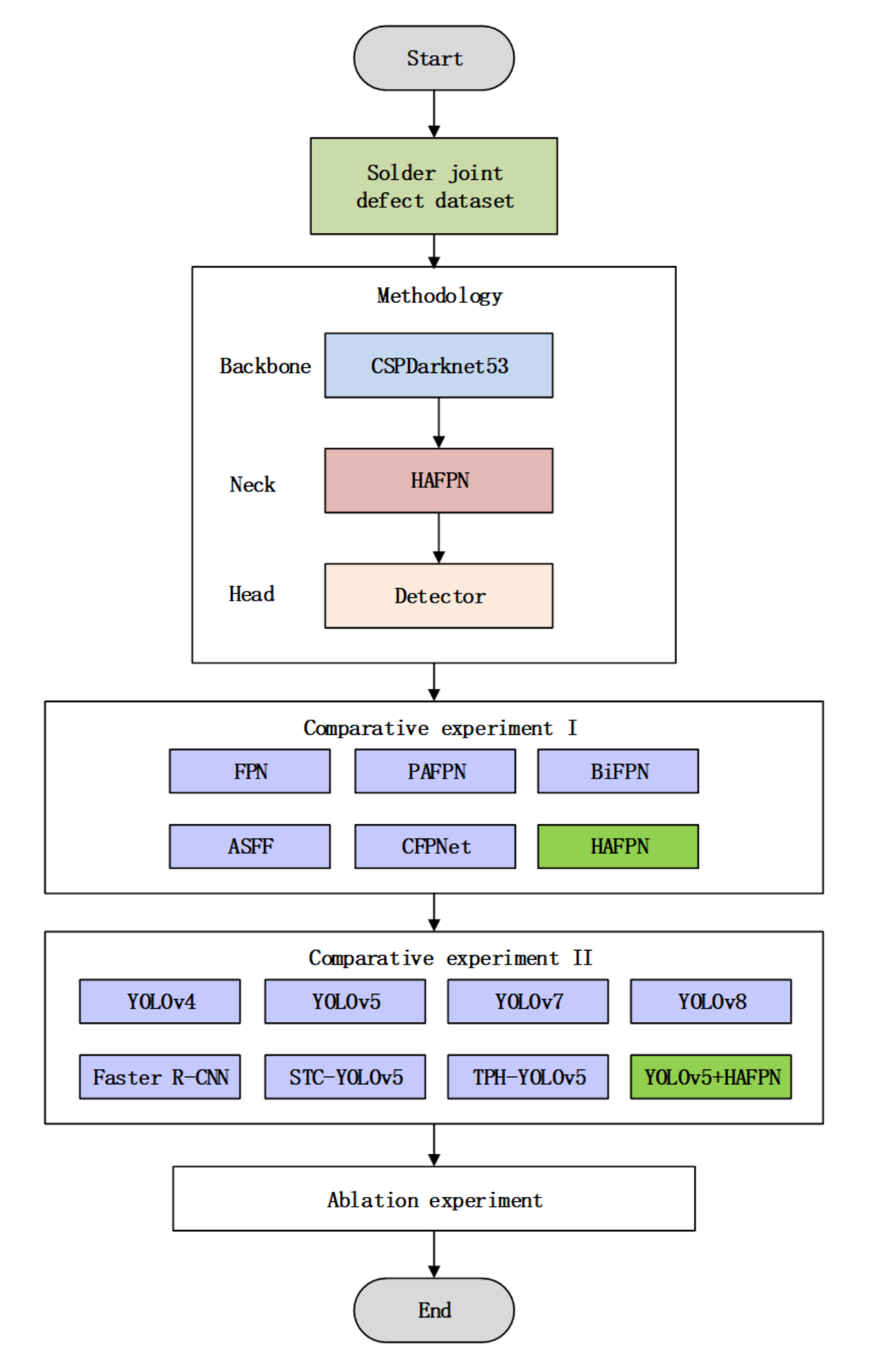}
			\caption{\label{fig:2} The overall process flowchart.}
		\end{figure}
		
		The main work and innovative points of this paper are as follows.
		
		(1) We propose a novel enhanced multi-head self-attention mechanism (EMSA) to enhance the ability of the network to perceive contextual information, improve the network utilization range of features, and enable the network to have more robust nonlinear expression capabilities.
		
		(2) We combine a coordinate attention mechanism (CA) with the EMSA to design a hybrid attention mechanism (HAM) network, which solves the problem of shallow feature loss in feature pyramid networks, increases the capacity of the network to perceive long-distance position information and learn local features.
		
		(3) The hybrid attention mechanism improves FPN and improves its ability to fuse functions and information transfer between network channels.
		
		(4) The improved FPN is applied to the YOLOv5 detection model, which improves the solder joint defect detection ability of YOLOv5, significantly solving the low detection rate issue of small defects, while enhancing the universal applicability of the defect detection model.

		\section{RELATED WORK}
		\subsection{Feature Pyramid Network}
		
		Feature Pyramid Network (FPN) \cite{11} is a feature fusion method commonly used for object detection, which is a network model for extracting pyramid feature representations. It is usually used in the feature fusion stage of object detection. After performing a bottom-up feature extraction operation on the backbone network, the FPN is connected to the corresponding layer's feature maps from top to bottom and horizontally, sequentially combining the two adjacent layers in the backbone network's feature hierarchy to construct a feature pyramid. Although FPN is simple and effective, some aspects still have shortcomings. Before the feature fusion at each layer, there is a semantic gap between different layers, and direct fusion will have a negative impact on the representation ability of multi-scale features. During feature fusion, the feature information at the high-level of the pyramid network will be lost during the upsampling process.
		
		Path Aggregation Network (PANet) \cite{12} structure has been improved based on FPN and is extensively employed in the YOLO object detection frameworks and its variant frameworks. This network has two feature fusion paths, namely top-down and bottom-up. This approach reduces the fusion distance between deep and shallow features. optimize the feature fusion method of FPN network to a certain extent, improve the object detection effect. However, due to the addition of a bottom-up path, low-level feature information will still be lost due to the deepening of network layers, and the additional paths increase computational complexity and network parameters, reducing the detection speed of the network model \cite{17,22}.
		Bi-directional Feature Pyramid Network (BIFPN) [13] introduces jump connections, which uses the jump connections to transfer information between input and output layers of features. Because the operation is in the same layer, this method can combine more features with fewer parameters. In order to accomplish more feature fusion, BIFPN calculates the same layer parameters more than once, treating each two-way path as one feature network layer.
		
		Adaptive Spatial Feature Fusion (ASFF) \cite{14} was proposed in 2019 as a feature fusion algorithm with adaptive capabilities. It can adaptively obtain important information through weight selection, improving the effectiveness of feature fusion. Being able to solve the inconsistency problem between features of different sizes in the feature pyramid by learning the connections between different feature maps. It has the advantages of easy implementation, cheap computational, and wide applicability.
		Quan et al. \cite{15} proposed a Centralized Feature Pyramid (CFP), it is based on global explicit centralized feature rules and can be used in object detection models. This scheme proposes a generalized intra layer feature adjustment method that uses lightweight multi-layer perceptron (MLP) to capture full length distance correlations, and emphasizes the use of intra layer feature rules, which can effectively obtain comprehensive but differentiated feature representations. CFP network can effectively improve the object detection capabilities of YOLOv5 and YOLOX. It improved mAP values by 1.4\% on the public dataset MS-COCO, but it’s computational complexity is relatively high.
		
		FPN is commonly employed in several instances involving defect detection. Chen et al. \cite{16} used YOLOv3 for SMD LED chip defect detection, using basic FPN as a feature fusion module. It has a reasonable detection rate for missing components, missing wire, and reverse polarity defects but a low detection rate for Surface defects. The reason is that the size of the surfaces defect is relatively small and the distribution position is uncertain, so it is difficult to detect. Yang et al. \cite{17} used YOLOv5 for steel surface defect detection, using Path Aggregation Feature Pyramid Network (PAFPN) as a feature fusion module to detect six types of defects on the steel surface, which achieved good real-time detection results, but had a low detection rate for small defect targets. The Precision value is 0.752, and the mAP value is 0.827. Du et al. \cite{18} used enhanced YOLOv5 for PCB defect detection, using BiFPN as a feature fusion module to detect surface defects on PCBs. The mAP50 index reached 95.3\%, but the mAP value was lower for the smaller defects of the mission hole and open circuit. The mission hole defect is the hole effect formed in the solder pad socket on the PCB due to a lack of solder. The Open Circuit defect refers to the defect where the circuit on the PCB is disconnected.
		
		Han et al. \cite{27} designed a YOLO improvement scheme that replaces the original PAFPN with BiFPN, and uses the self-attention mechanism to embed the up sampling and down sampling processing modules in BiFPN, improving the detection rate of the model in surface defect detection tasks. However, the ability to detect smaller defects is weak. Therefore, in order to improve the detection performance of defect detection networks, it is necessary to design an enhanced attention mechanism to improve the feature fusion ability of FPN, thereby reducing the missed detection rate of small-sized defects.
		In recent years, many studies have utilized attention mechanisms to enhance the detection capabilities of defect detection frameworks. Attention mechanism is a mechanism that enables neural networks to focus their attention on a specific object. 
		
		\subsection{Attention Mechanism}
		The numerous input information includes critical and irrelevant information required by the task. The attention mechanism can focus on these key information while filtering irrelevant information. The inspiration for the attention mechanism comes from the vision system in humans, which can quickly browse images, locate the target area of interest, and enhance attention to the target area, thereby obtaining important information in that area and suppressing interference from other unrelated areas. Hu et al. proposed an attention module called Squeeze and Stimulation (SE) \cite{19}. This attention module adaptively corrects the weight parameters of each channel by mining the inter-dependencies between feature channels so that the network can pay attention to more critical feature information. Woo et al. \cite{20} extended the spatial dimension and designed the Convolutional Block Attention Module (CBAM). Its sequential construction of the Channel Attention Module (CAM) and the Spatial Attention Module (SAM) enhances the network's capacity to separate and reinforce feature information. The Efficient Channel Attention (ECA) module \cite{21} uses one-dimensional convolutional operations to extract dependency relationships between channels, achieving cross channel interaction. It solves the problem of SE's insufficient extraction of dependency relationships between channels due to compression dimensionality reduction. ECA has a low computational complexity and has little impact on the speed of the network.
		
		Zhang et al. \cite{22} embedded ECA into the feature fusion network in YOLOv5 for solar cell surface defect detection, enhancing PAFPN's ability to fuse solar cell surface defect features and further improving the defect detection rate. The mAP50 value on the dataset reached 84.23\%. However, ECA incurs significant computational overhead for smaller feature maps. In order to better detect surface defects on steel, Qian et al. \cite{23} introduced the CA mechanism into the detection network. The mAP value is 79.23\%, while the Recall value is only 62.4\%. The CA mechanism requires the calculation of attention weights for the entire feature map, so it cannot capture long-distance dependencies. Gathering semantic information for long-distance dependencies is crucial for small-area detection. On the other side, the Vision Transformer (ViT) \cite{24} relies entirely on self-attention to capture long-distance global relationships and has better accuracy than Convolutional Neural Network (CNN). ViT was introduced into computer vision in 2020, and much research has proven its performance in the field of vision.
		
		\subsection{Vision Transformer}
		Vision Transformer has achieved good performance in the field of computer vision because it uses the multi-head self-attention (MSA) mechanism. The MSA mechanism is a feature extraction method different from CNN, which can establish global dependencies and expand the receptive field of images. Compared to CNN, Its receptive area is larger, and it can gather more contextual information. However, some vital information needed for the detection is removed due to inefficient filtering. Prior knowledge of feature localization, translation invariance, and image scale is not utilized. The ViT ability to capture adequate information is weaker than CNN, and it cannot utilize the prior knowledge of feature localization, translation invariance, and image scale of the image itself. The model design of ViT adopts a scaling dot product attention mechanism. ViT first divides the image into non-overlapping, fixed-size image blocks and flattens the image blocks into one-dimensional vectors for linear projection to achieve feature extraction. 
		
		Another type of Transformer is the Swin Transformer. Swin Transformer \cite{25} utilizes local attention and the displacement window multi-head self-attention mechanism (SW-MSA) to achieve interaction between local and global features, achieving good results in various visual tasks and solving the problem of ViT local information being easily damaged.
		
		The difference between the self-attention mechanism and the attention mechanism is that the queries and keys come from different sources, while the queries and keys of the self-attention mechanism come from the same set of elements. Zhu et al. \cite{26} designed a Transformer Prediction Head YOLOv5 (TPH-YOLOv5) model for tiny object detection in drone images. The model uses the the Transformer to detect low-resolution feature maps, enhancing the network's capability to extract different local information and achieving better performance for high-density objects. However, using the Transformer module in multiple parts of the model resulted in a significant computational workload. 
		
		\section{PROPOSED ENHANCED FEATURE PYRAMID NETWORK}
		
		\subsection{Hybrid Attention Feature Pyramid Network Architecture}
		
		In the task of detecting solder joint defects, some small defects are challenging to detect. Strengthening the feature fusion ability of FPN can help improve the detection effect of small defects. To enhance the feature fusion ability of FPN, this study proposes a Hybrid Attention Feature Pyramid Network (HA-FPN) shown in Figure 3(a). Adding a Hybrid Attention Mechanism (HAM) in the basic FPN enhances FPN's ability to perceive contextual information. It also expands its utilization of feature information and solves the problem of severe loss of location information. The HAM network structure is shown in Figure 3(b).
		
		\begin{figure}[!t]
			\centering
			\subfigure[]{
				\includegraphics[scale=0.3]{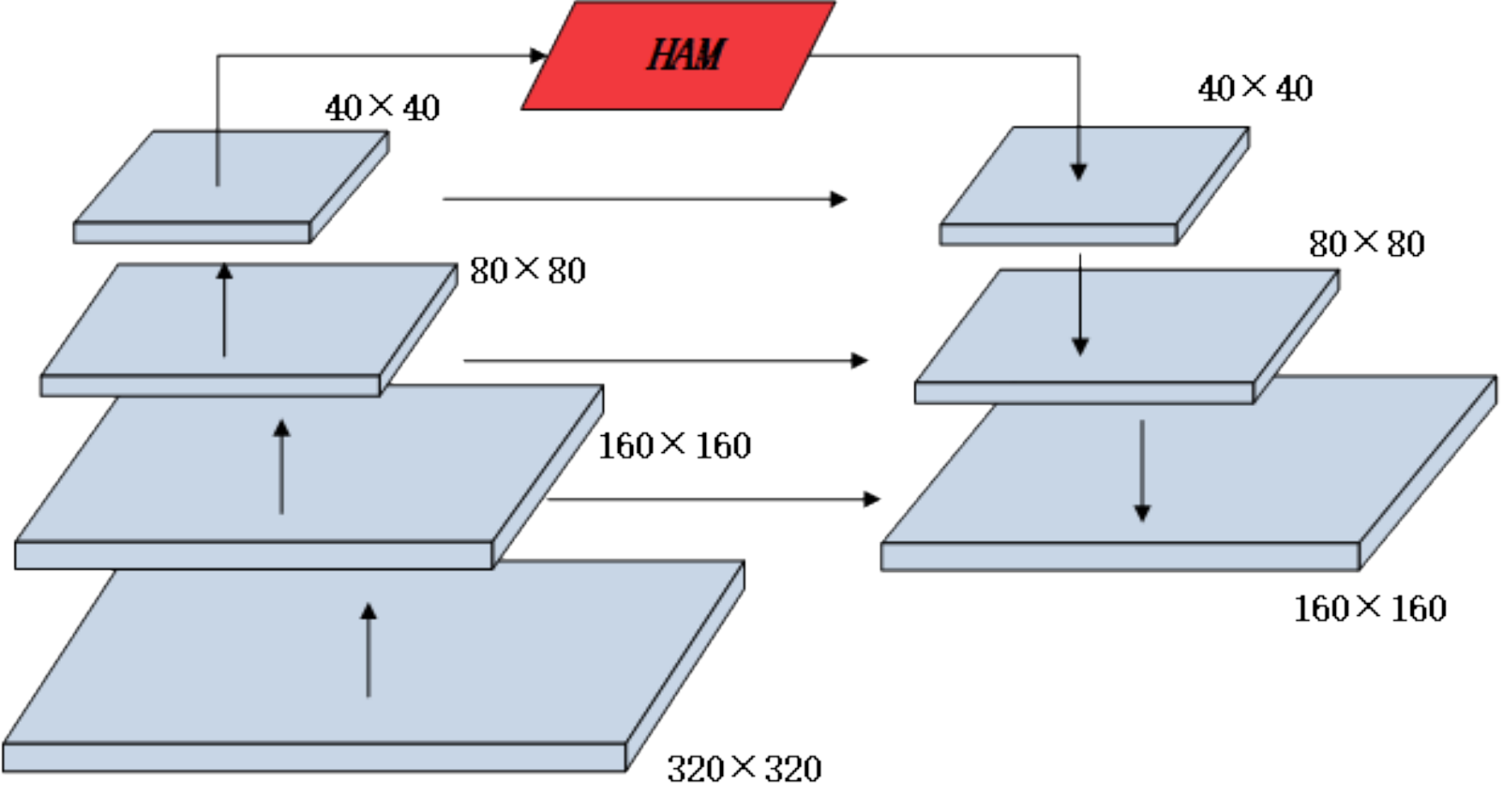}}
			\subfigure[]{
				\includegraphics[scale=0.25]{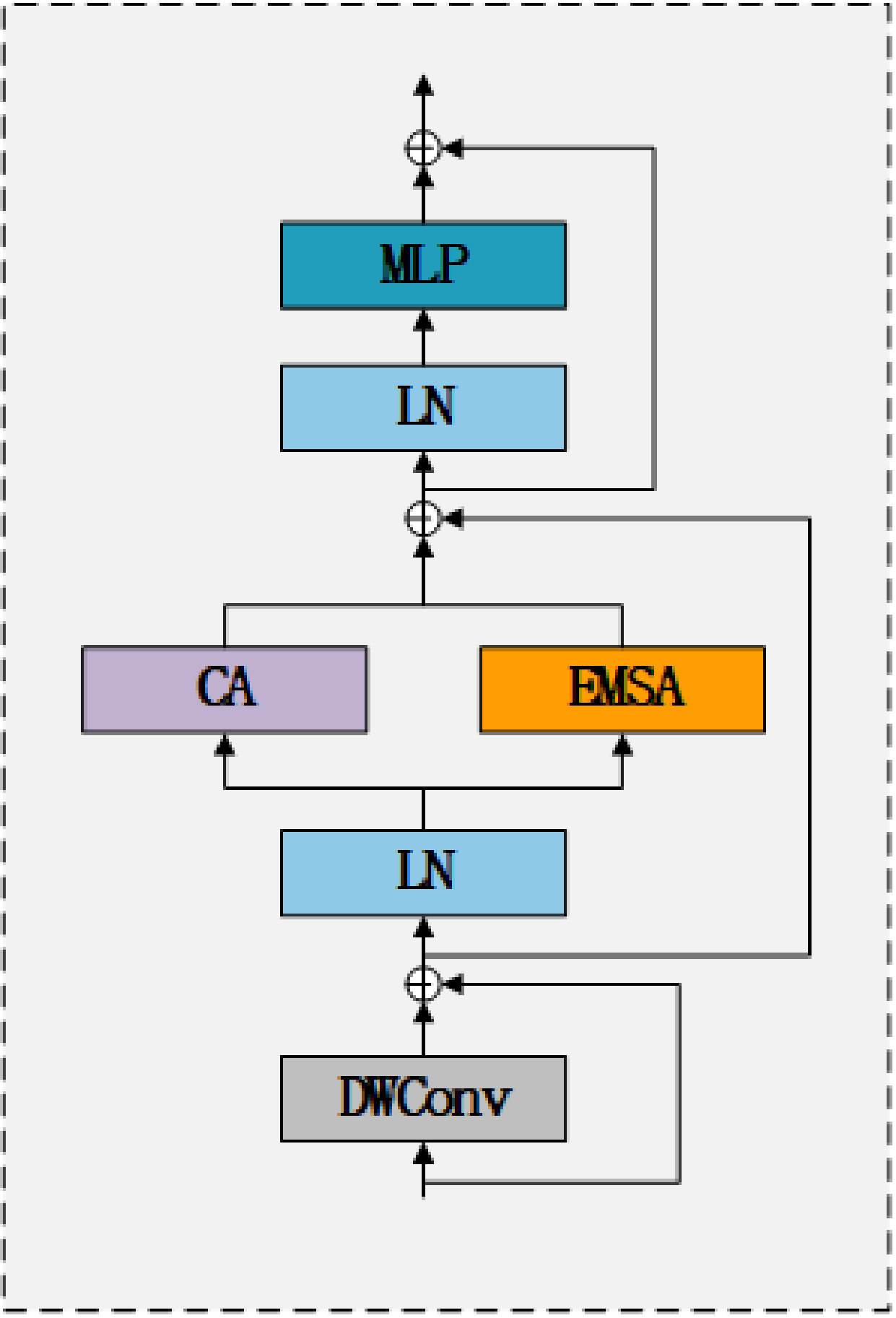}}
			\caption{Hybrid Attention Feature Network Architecture. (a)Hybrid Attention Feature Network (HAFPN), (b)Hybrid Attention Mechanism (HAM)}
			\label{fig_3}
		\end{figure}

		\subsection{Hybrid attention mechanism}
		
		The Hybrid Attention Mechanism (HAM) module is based on the Transformer structure.  Firstly, the input features are passed through a Depth Wise Convolution (DWConv) residual block to achieve parameter sharing and enhance the learning of local features. Next, Layer Normalization (LN) is used for normalization processing. The output is processed through two attention mechanism modules, Enhanced Multi Head Attention (EMSA) and CA. After processing, it is normalized through an LN layer, and finally, the processing results are output through the MLP layer. The entire processing process is shown in Equation (3.1).
		
		\begin{equation}\label{eqn-1}
			\begin{aligned}
				& X_1=X+D W \operatorname{conv}(X) \\
				& X_2=\operatorname{LN}\left(X_1\right) \\
				& X_3=C A\left(X_2\right)+\operatorname{EMSA}\left(X_2\right)+X_1 \\
				& Y=\operatorname{MLP}\left(\operatorname{LN}\left(X_3\right)\right)+X_3
			\end{aligned}
		\end{equation}
		
		In the 	Equation (3.1), X represents input features, Y represents output features, X1, X2, and X3 are intermediate features. DWconv represents deep separable convolution, LN represents layer normalization, CA represents coordinate attention, and EMSA represents enhanced multi head self-attention. MLP is Multilayer Perceptron.
		
		(1)Enhanced Multi-head Self Attention
		
		A novel EMSA module, as shown in Figure 3 (b), has been proposed for obtaining contextual information and global features, simultaneously using the CA mechanism to capture accurate positional features and capture information between channels effectively. Then, the fusion of information features captured by EMSA and CA to enhance the feature pyramid network's feature fusion capability is executed. The design concept is based on the MSA mechanism in Transformer, as shown in Figure 4 (a). 
		
		\begin{figure}[!t]
			\centering
			\subfigure[]{
				\includegraphics[scale=0.3]{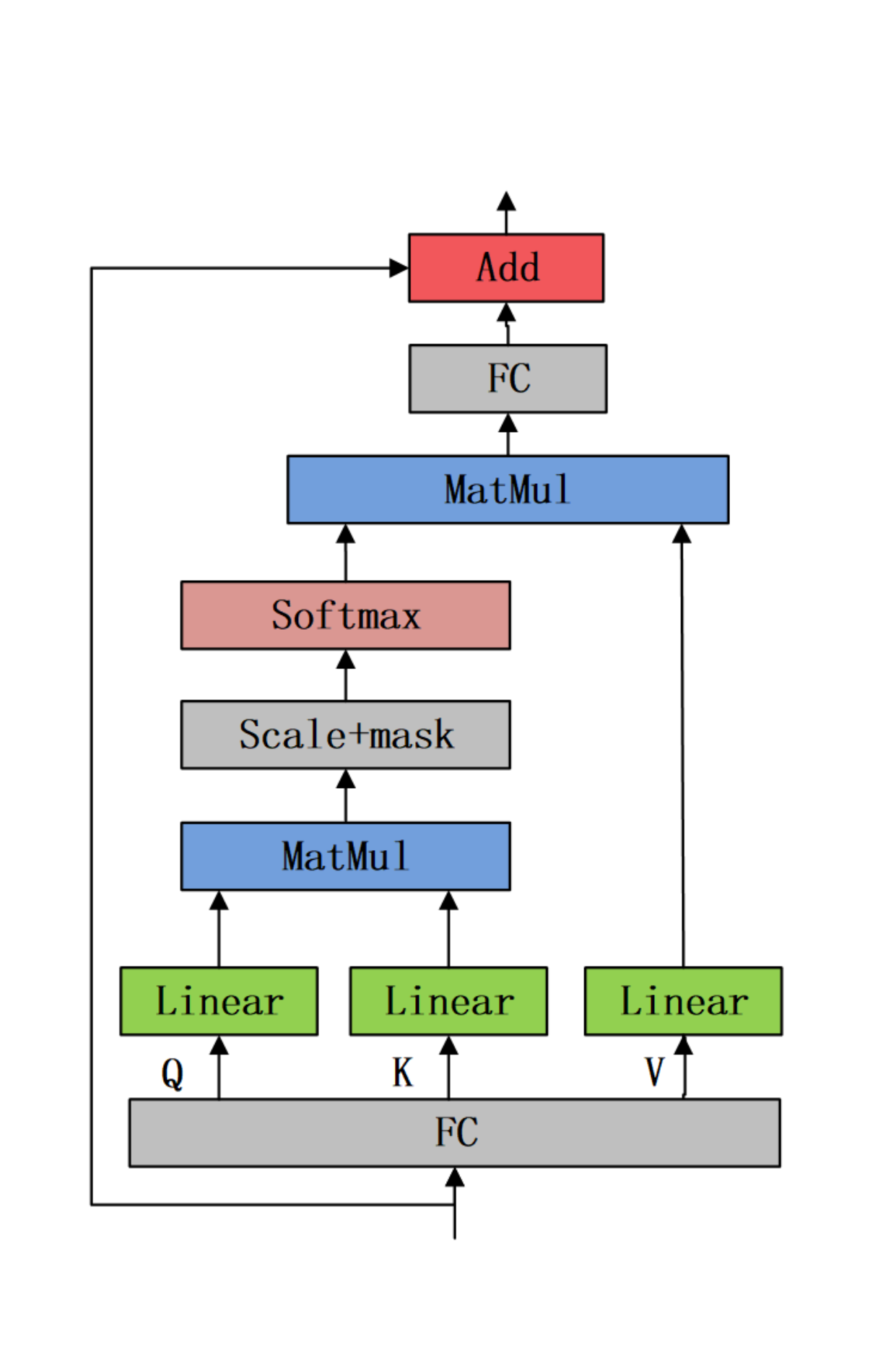}}
			\subfigure[]{
				\includegraphics[scale=0.3]{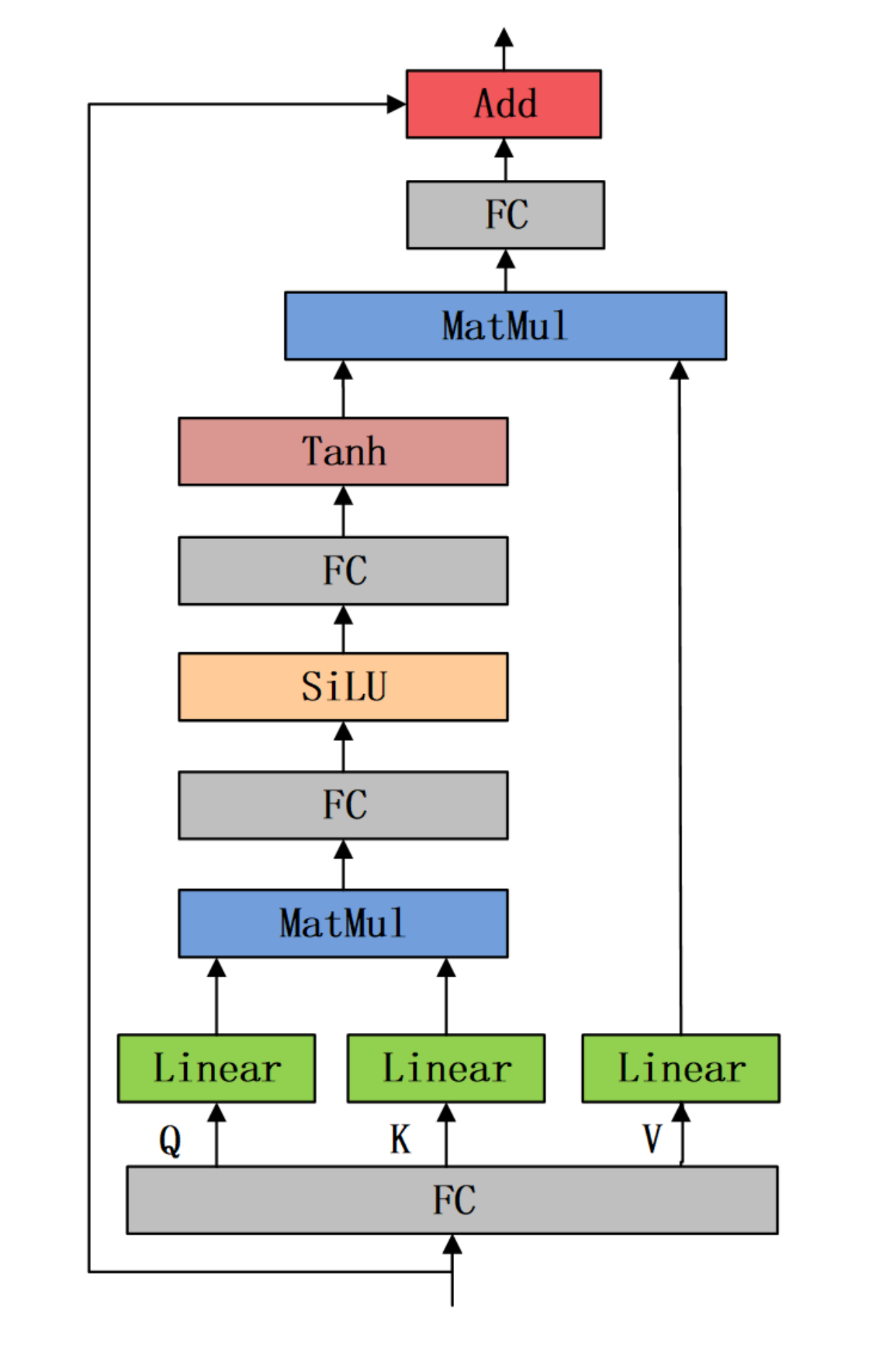}}
			\caption{Hybrid Attention Feature Network Architecture. (a)Muti-Head Self Attention, (b)Enhanced Muti-Head Self Attention(EMSA)}
			\label{fig_3}
		\end{figure}
		
		The architecture of EMSA is shown in Figure 4 (b). The entire processing process of EMSA is shown in Equation (3.2). 
		
		\begin{equation}
			\begin{aligned}
				& Q, K, V=F C\left(X_{\text {input }}\right) \\
				& Q^{\prime}=\operatorname{Linear}(Q) \\
				& K^{\prime}=\operatorname{Linear}(K) \\
				& V^{\prime}=\operatorname{Linear}(V) \\
				& X_{\mathrm{m}}=\operatorname{SiLU}\left(F C\left(Q^{\prime} \otimes K^{\prime}\right)\right) \\
				& X_{\mathrm{n}}=\operatorname{Tanh}\left(F C\left(X_{\mathrm{m}}\right) / \sqrt{d}\right) \\
				& X_{\text {output }}=F C\left(X_{\mathrm{n}} \otimes V^{\prime}\right)+X_{\text {input }}
			\end{aligned}
		\end{equation}
		
		In the Equation (3.2), $X_{\text {input }}$represents input features, $X_{\text {output }}$represents output features, $X_{\mathrm{m}}$ and $X_{\mathrm{n}}$ represent intermediate features. Q, K and V represent the query matrix, key matrix, and value matrix, respectively. Linear is a linear transformation operation, SiLU is the Sigmoid Linear Unit activation function, FC represents Full Connection processing, and d is the scalar factor.
		
		Firstly, Q, K, and V components are formed through a fully connected (FC) layer, and then linear transformations are performed on each of the three components. Multiply the transformed Q and K matrices, and then perform a series of nonlinear treatments on them. Then, a fully connected layer is used to input the Silu activation function, which is very similar to the model of signal transmission within neurons, thus more in line with some biological implementation mechanisms and better simulating the information processing mechanisms of the human brain. After passing through a fully connected layer, the Tanh activation function is used for processing, and the output result is a matrix multiplied with the linearly transformed V component. Finally, an FC layer is fused with the original input features to obtain the final output result. Compared to the vanilla MSA, the EMSA has more nonlinear transformations, which can make the attention network have stronger context awareness ability, further enhancing the network's utilization range of features, and making the network have stronger nonlinear expression ability. 
		
		(2)Coordinate attention
		
		This study introduces the coordinate attention (CA) mechanism \cite{28} into HAM to enhance the fusion capability of FPN for location information. The CA mechanism can effectively enhance the association between different channels and improve the network's perception ability for long-distance location information. The operation process of the CA mechanism is shown in Figure 5. For input H (Height of the input feature map) $\times$W (Weight of the input feature map) $\times$C (Channel of the input feature map), Firstly, global average pooling is performed from both the height and width dimensions of the image to obtain feature maps with sizes H$\times$1$\times$C and 1$\times$W$\times$C; Next, concatenate the feature maps of two sizes and reduce the dimensionality from the channel dimension through shared convolution to obtain a feature map of size $1 \times(\mathrm{W}+\mathrm{H}) \times \mathrm{C} / \mathrm{r}$. After passing through the nonlinear layer, its nonlinear expression ability is improved. Subsequently, In order to increase the dimensionality, 1$\times$1 convolution is utilized, restoring the feature maps from the width and height dimensions to the A and B scales, and assigning weights through HardSigmaid. To accelerate the processing speed of the CA mechanism, HardSigmoid is used to replace the original Sigmoid activation function for weight assignment. HardSigmoid does not require exponentiation, so its calculation speed is faster than Sigmoid. Finally, The feature map's size is changed to H$\times$W$\times$C.
		
		\begin{figure}
			\centering
			\includegraphics[width=0.6\linewidth]{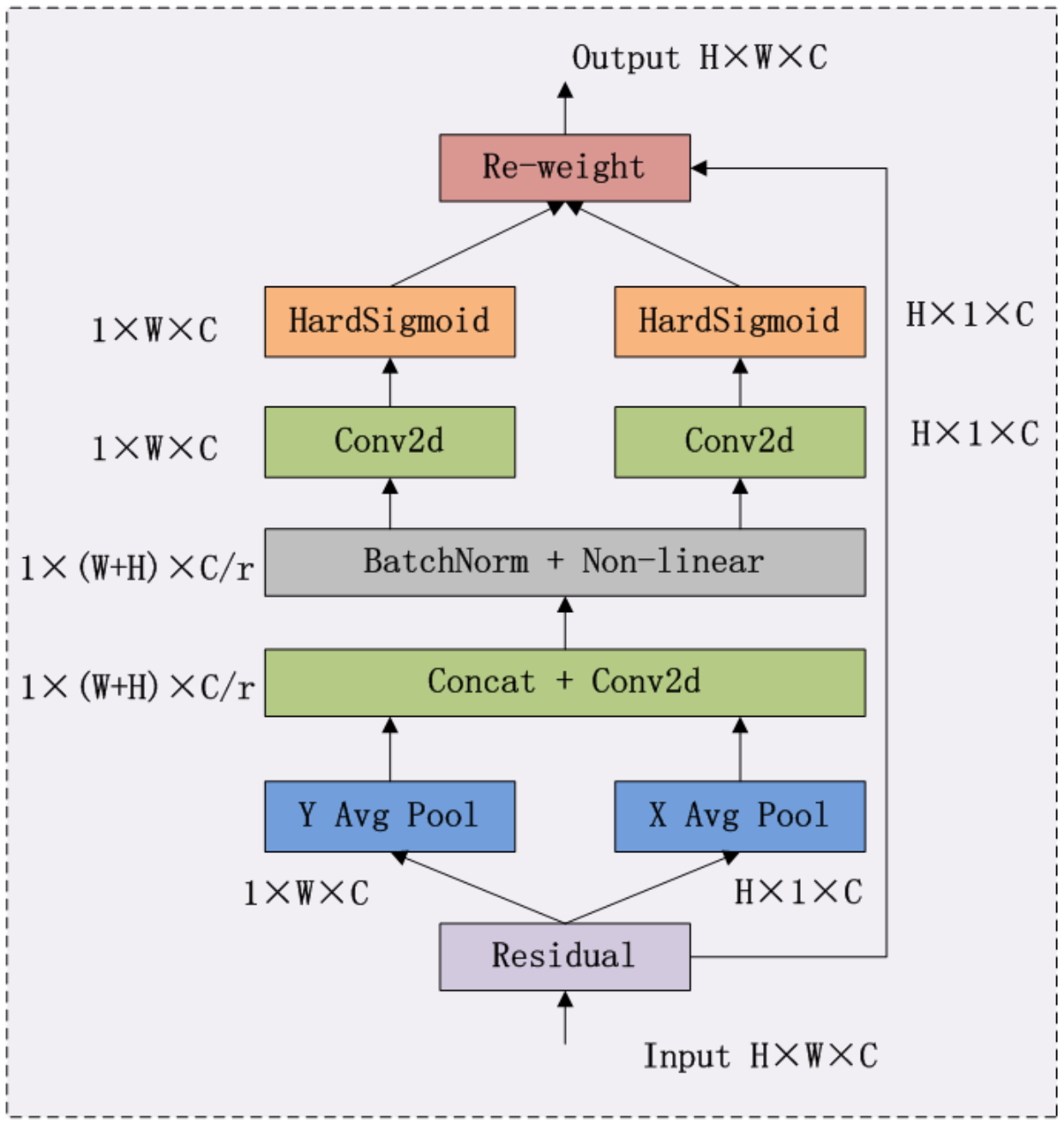}
			\caption{\label{fig:5} Coordinate Attention Mechanism.}
		\end{figure}

		\subsection{Improved Feature Fusion Network In YOLOv5}
		We use the HAFPN as a feature fusion module in YOLOv5, replacing the original PAFPN structure. The original feature fusion network architecture is shown in Figure 6(a). It comprises of Convolution + Batch Normalization + SiLu activation function (CBS), Cross Stage Partial (CSP) Bottleneck with 3 convolutions (C3) and  Spatial Pyramid Pooling Fast (SPPF). Compared to FPN, PAFPN has better network accuracy, but its detection effect for some small defects in solder joints is not good, and the network size is large and has many parameters. Our proposed method strengthens the feature fusion ability of the FPN network, striving to improve recognition accuracy while ensuring detection speed. The original feature fusion network architecture is shown in Figure 6(b).
		
		\begin{figure}
			\centering
			\includegraphics[width=0.9\linewidth]{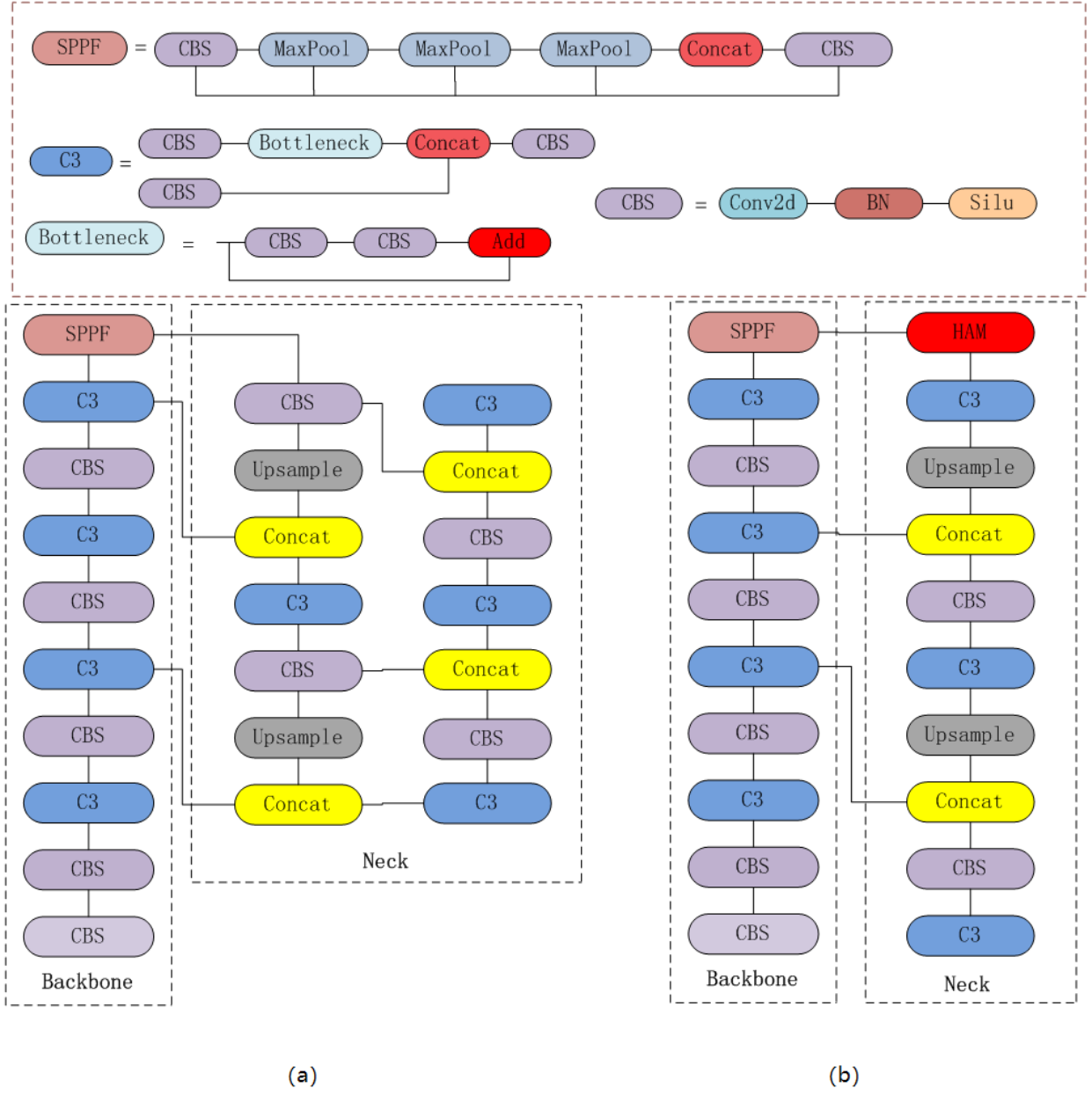}
			\caption{\label{fig:6} Original and Improved Feature Fusion Network in YOLOv5. (a) Original Feature Fusion Network in YOLOv5, (b) Improved Feature Fusion Network in YOLOv5
			}
		\end{figure}
		
		\section{EXPERIMENT}
		\subsection{Experimental Environment}
		The hardware and software environment of this experiment is shown in Table 1.
		
		\begin{table}
			\centering
			\fontsize{10}{15}\selectfont    %{字体尺寸}{行距}
			\caption{Experimental environment}
			\begin{tabular}{ll}
				\toprule
				Environment & Configuration/Version \\ \cmidrule(lr){1-2} 
				CPU              & Intel(R) Core™ i5-12400F 2.50GHz  \\ 
				GPU              & NVIDIA RTX 3060                  \\ 
				Video memory     & 12GB                             \\ 
				RAM              & 32 GB RAM                        \\ 
				PyTorch          & v1.13.1                          \\ 
				CUDA             & v11.3                            \\ 
				cudnn            & v8.0                             \\ 
				Operating System & Windows 10 Pro                   \\ 
				\bottomrule
			\end{tabular}
		\end{table}
		
		\subsection{Datasets}
		The solder joint defect dataset contained 3154 defective solder joint images obtained using a Couple-Charged Device (CCD) industrial camera. There are two types of defects, namely ineffective and foot shifting. Among them, 1680 are ineffective defects and 1474 are foot shifting defects. The dataset is randomly divided into training sets, validation sets, and test sets, with a division ratio of 80\%, 10\%, and 10\%, for model training, validation, and testing. The method of preparing the dataset is adapted and modified by \cite{17}. 
		
		\subsection{Evaluation criterion}
		The evaluation indicators for this experiment are Precision as calculated in formula (4.1), Recall as calculated in formula (4.2), Mean Average Precision (mAP) as calculated in formula (4.3), and Frames Per Second (FPS). In the formula, TP refers to the number of correctly predicted positive samples, FP refers to the number of incorrectly predicted positive samples, and FN refers to the number of predicted negative samples but actually positive samples. P represents Precision, and R represents Recall. 
		
		\begin{equation}{Precison}=\frac{TP}{TP+FP}\end{equation}
		
		\begin{equation}{Recall}=\frac{TP}{TP+FN}\end{equation}
		
		\begin{equation}mAP=\frac{\sum_{i=1}^NAP_i}{N} \qquad AP=\int_{0}^{1}P(R)dR\end{equation}
		
		\subsection{Experiments and analysis of results} 
		
		\hspace*{\fill} \\(1) Comparative experiment
		
		To verify the effect of hybrid attention proposed in this study, a heat map visualization was used to compare the focusing ability of different attention mechanisms on defect areas, as shown in Figure 7. When not utilizing attention mechanisms, YOLOv5 has a weaker ability to pay attention to solder joint defects. After adding several attention mechanisms, it shows some improvement. Among them, the attention mechanisms of SE and ECA have less improvement in defect attention ability and even have a declining effect. CBAM and CA attention have an enhancing effect on defect attention. Transformer and Swin Transformer have poor attention to shifting defects in which size is small. The hybrid attention proposed in this study significantly increases the coverage effect of the heat map at the defect location. It has a more vital ability to focus on small defects. The location positioning is more accurate, proving that hybrid attention can combine contextual content to focus on more pixels and also proving the effectiveness of hybrid attention.
		
		\begin{figure}[!t]
			\centering
			\subfigure[Original image]{
				\includegraphics[scale=0.16]{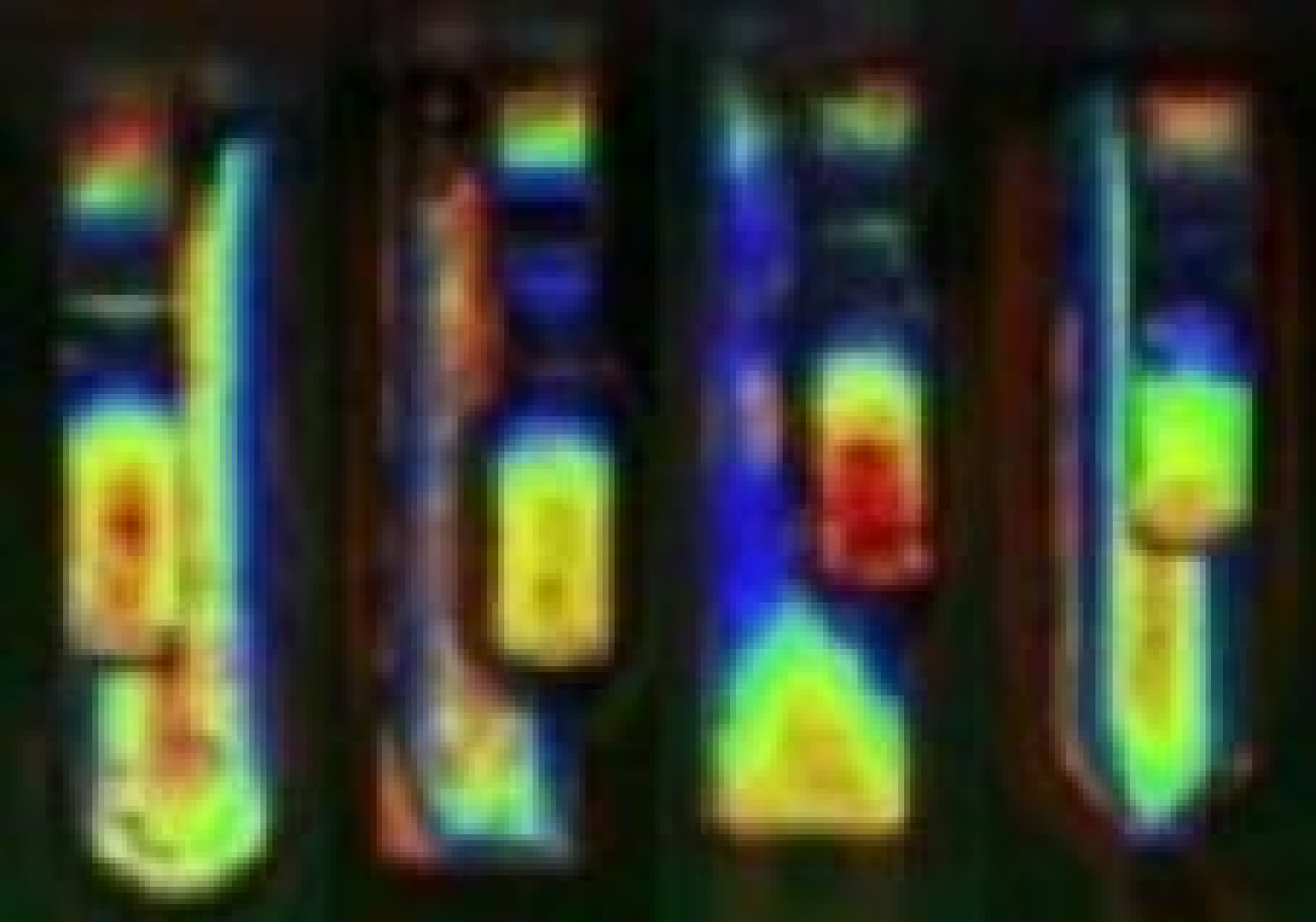}}
			\subfigure[Without attention mechanism]{
				\includegraphics[scale=0.16]{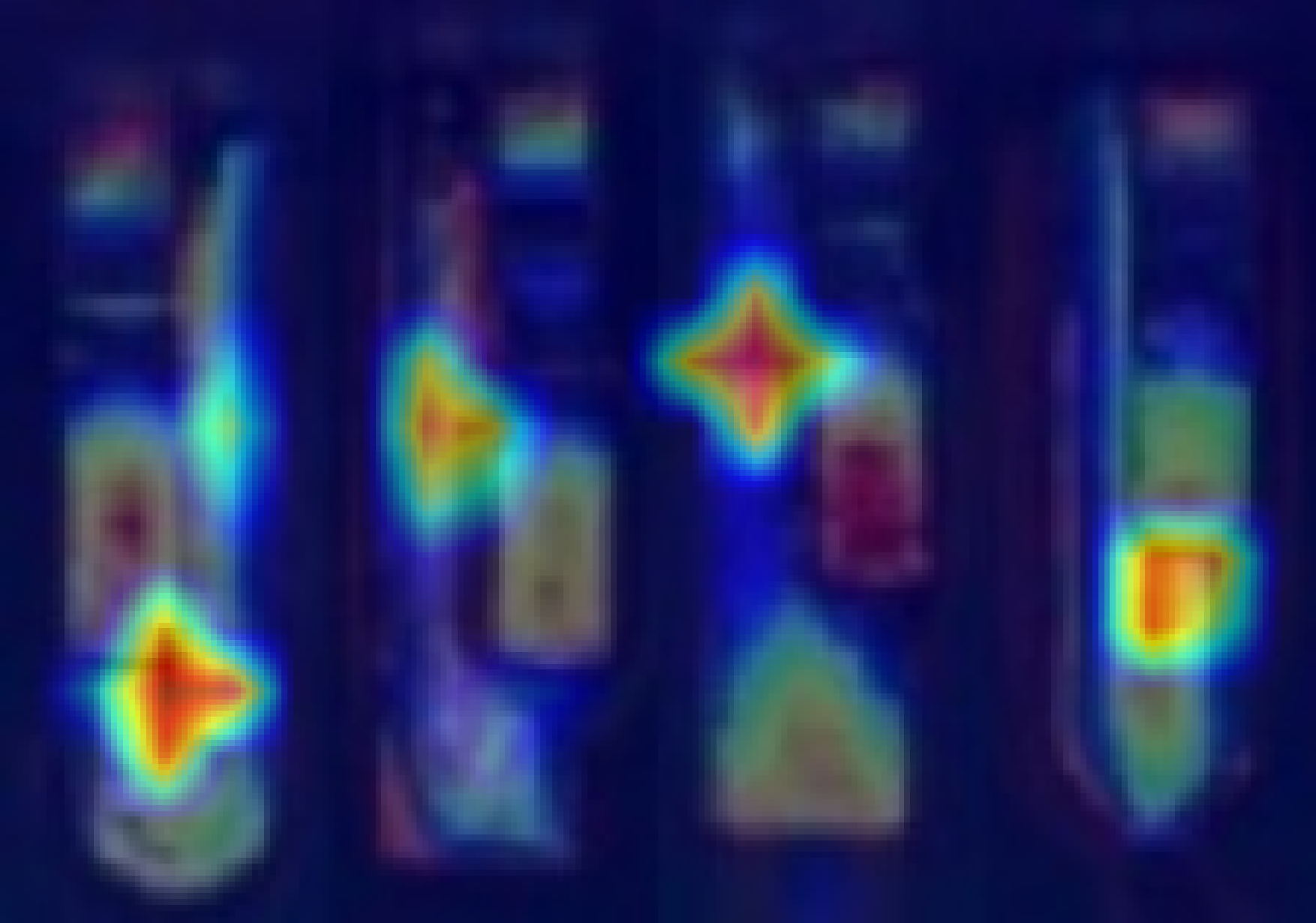}}
			\subfigure[Squeeze and Excitation]{
				\includegraphics[scale=0.16]{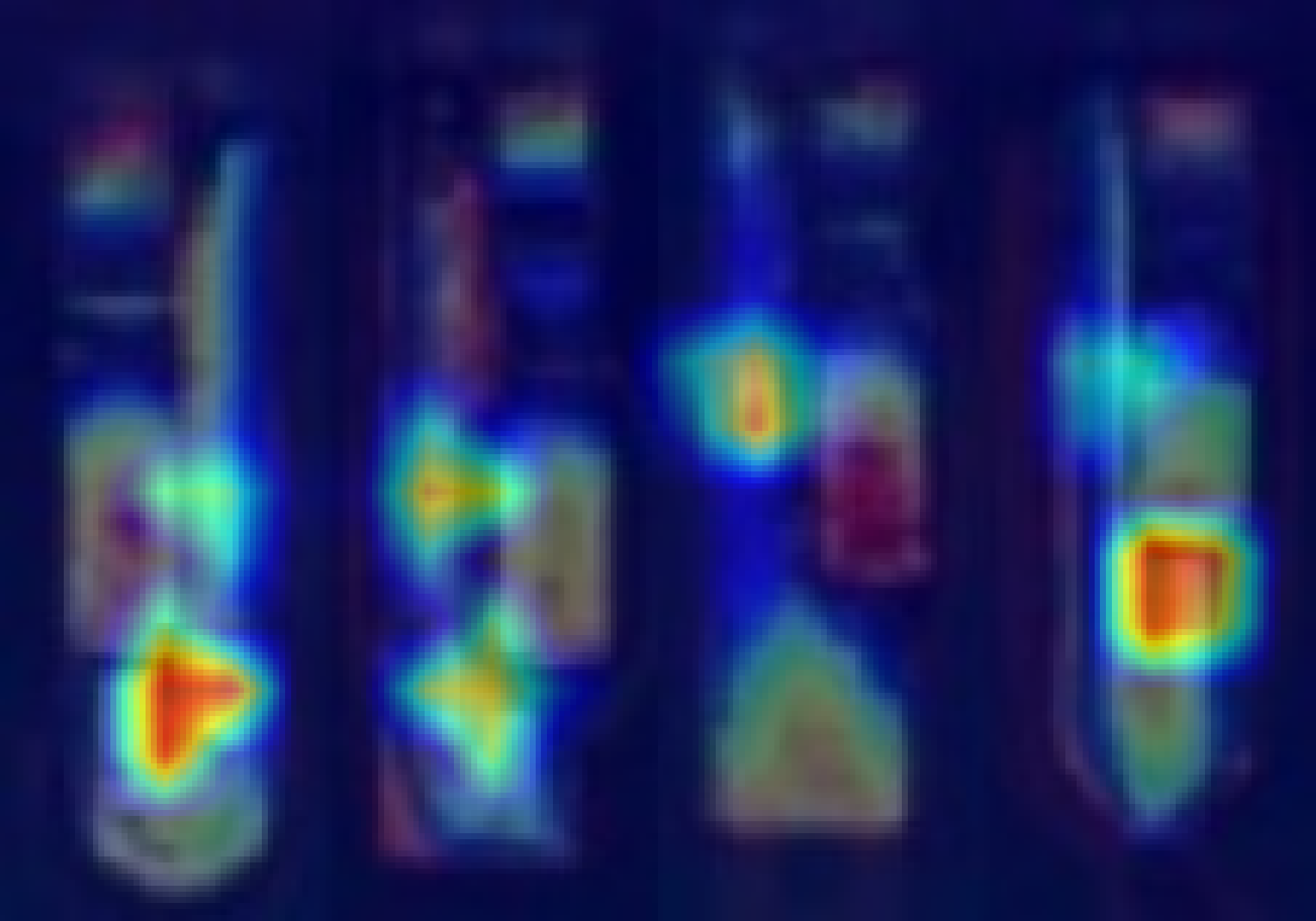}}
			\\
			\subfigure[CBAM]{
				\includegraphics[scale=0.16]{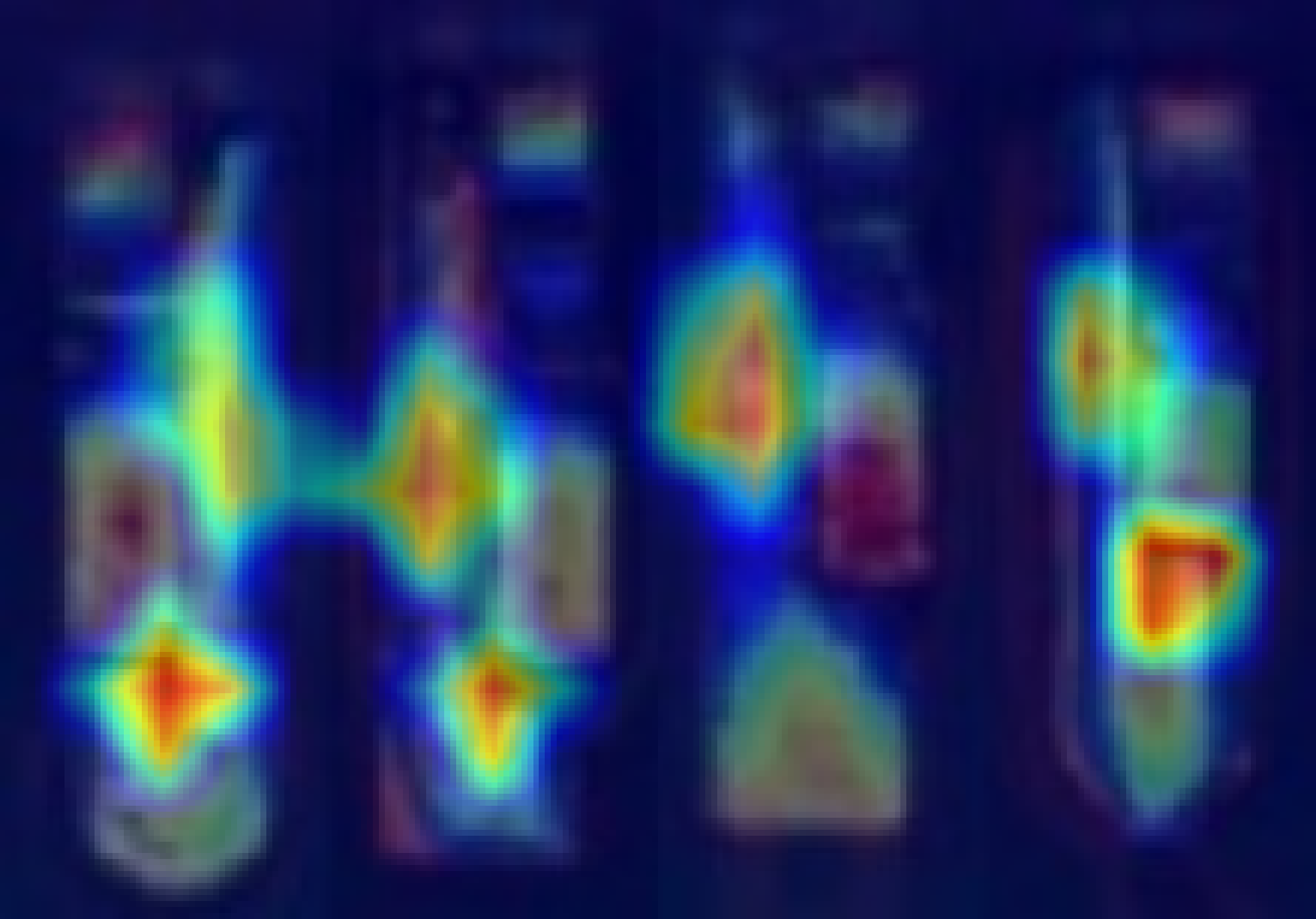}}
			\subfigure[Efficient Channel Attention]{
				\includegraphics[scale=0.16]{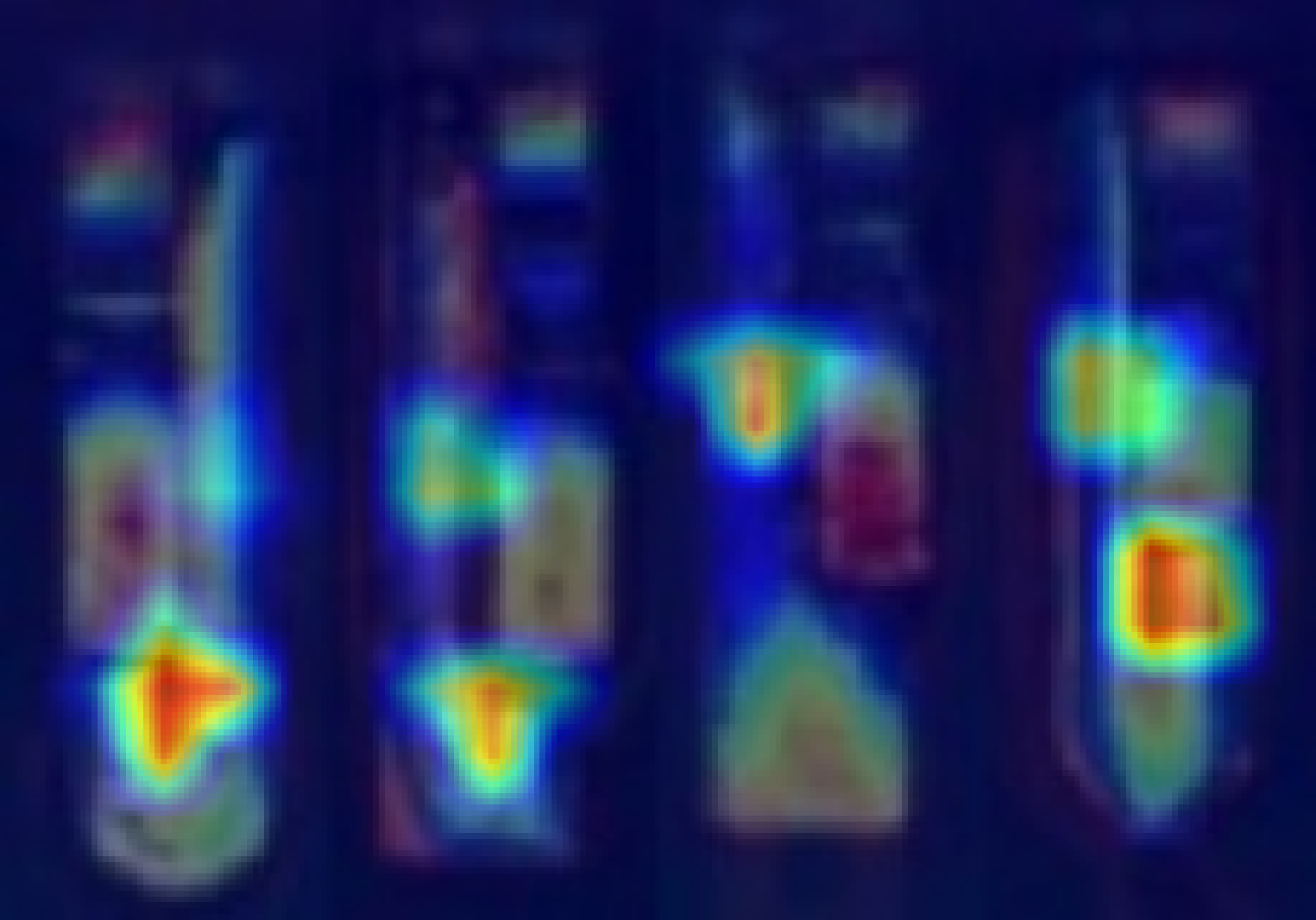}}
			\subfigure[Coordinate Attention]{
				\includegraphics[scale=0.16]{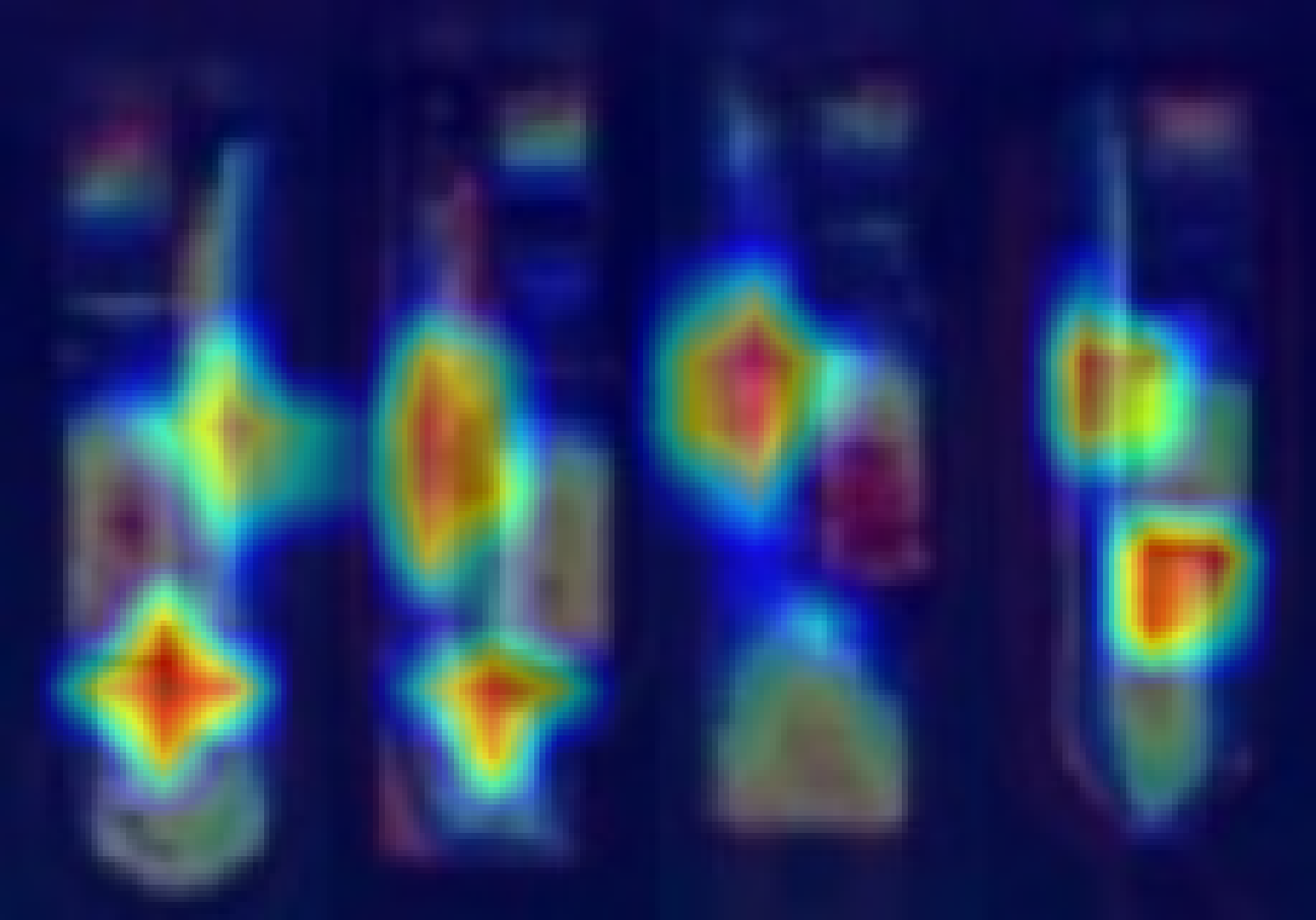}}
			\\
			\subfigure[Transformer]{
				\includegraphics[scale=0.16]{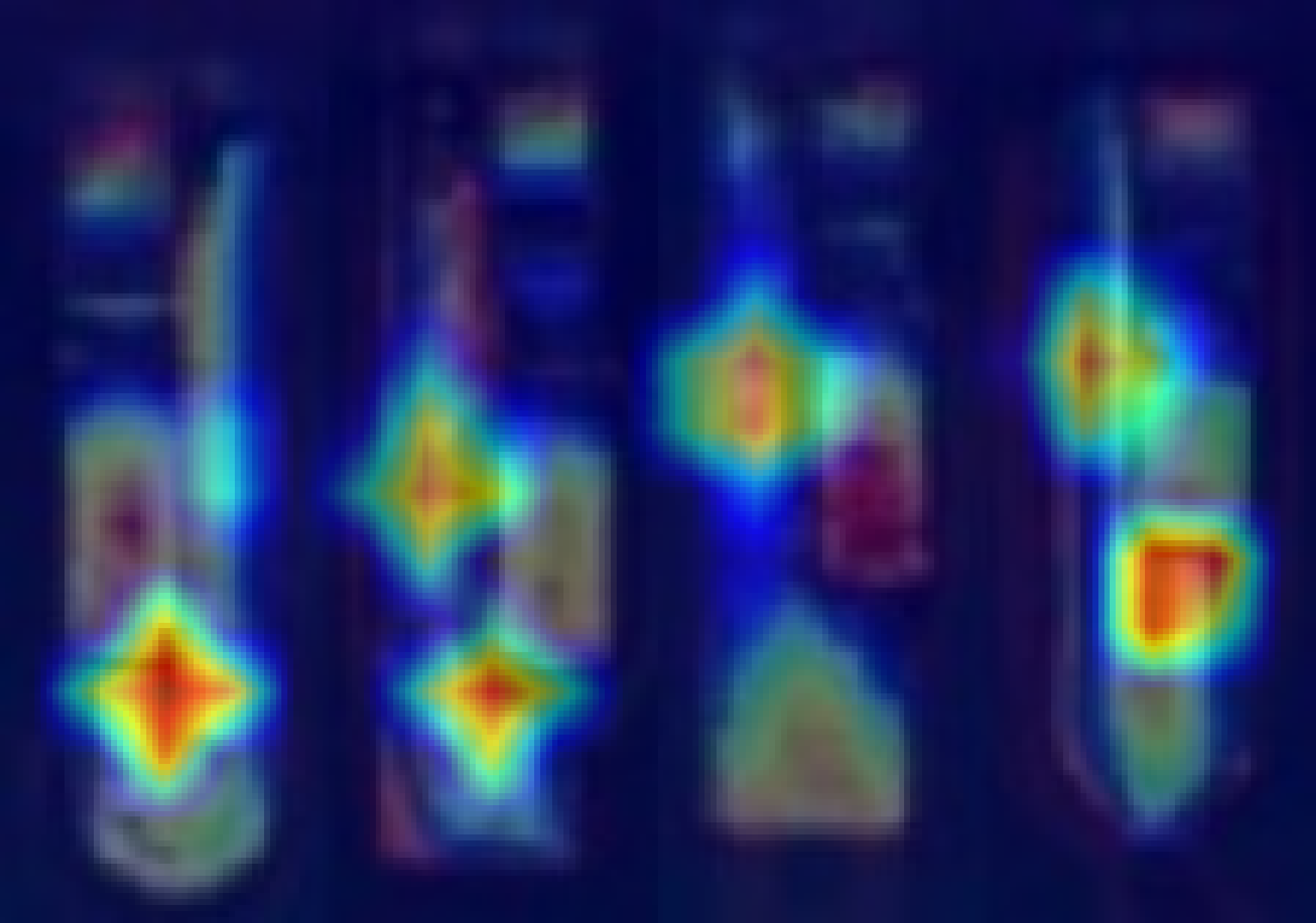}}
			\subfigure[Swin Transformer]{
				\includegraphics[scale=0.16]{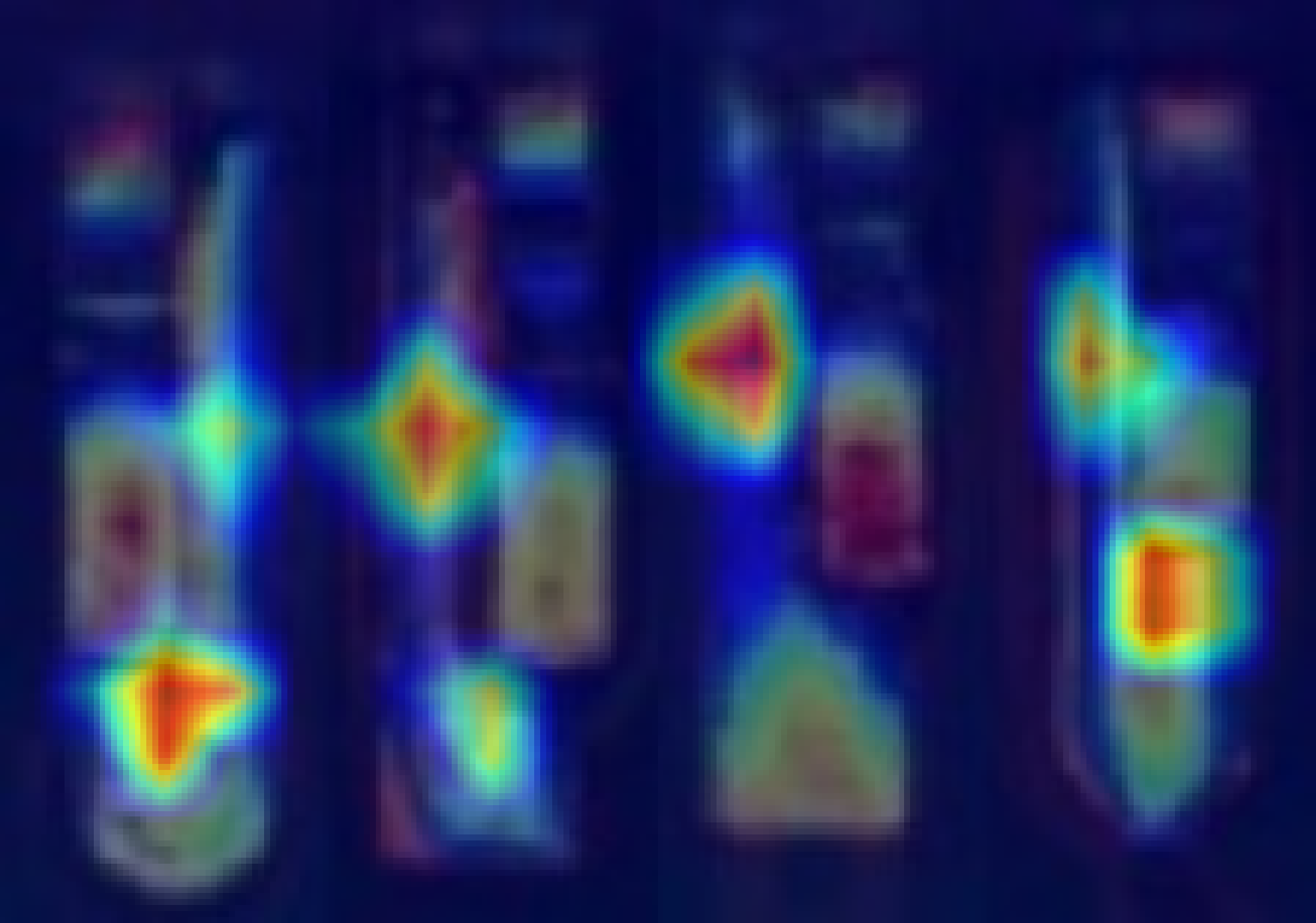}}
			\subfigure[Hybrid Attention Mechanism]{
				\includegraphics[scale=0.16]{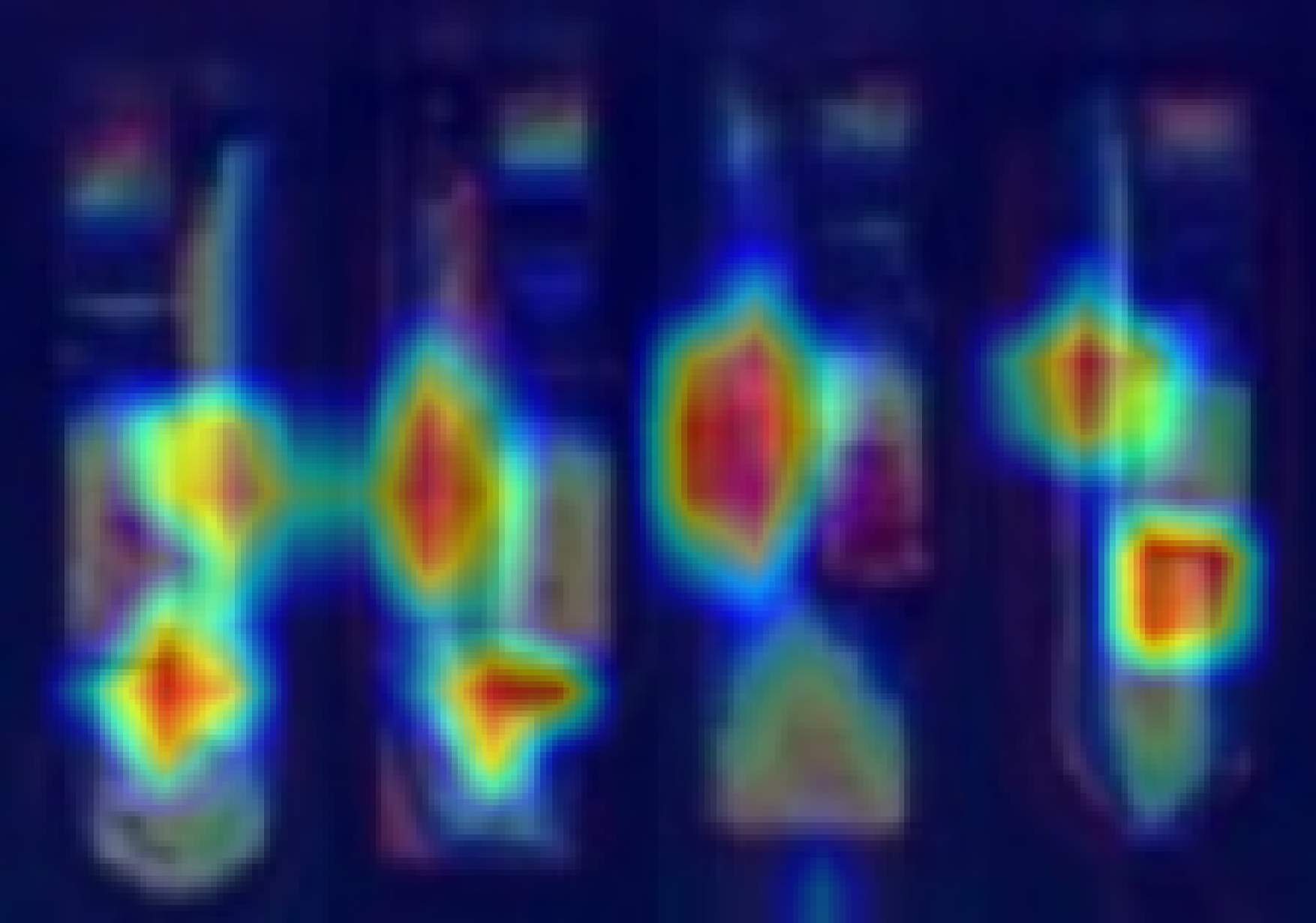}}
			\caption{Comparison of Heat Map Effects of Different Attention Mechanisms}
			\label{fig_7}
		\end{figure}

		To verify the superiority of the HAFPN algorithm, we compared the defect detection performance of different FPN algorithms on the same dataset. The CSPDarknet53 was used consistently as the feature extraction backbone network. The comparison feature fusion algorithms are FPN, PAFPN, ASFF, BiFPN, and CFPNet. The results are shown in Table 2. The detection indicators for all defects are higher than those of FPN, PAFPN, BiFPN, and CFPNet. The precision of Insufficient defects is slightly lower than ASFF. The overall precision, recall, and mAP values of HAFPN are superior to other networks, with precision being 3.8\%, 9.4\%, 1.3\%, 9.7\%, 6.9\% higher, recall being 0.5\%, 4.8\%, 0.7\%, 1.5\%, 1.2\% higher, and mAP being 3\%, 4.3\%, 0.9\%, 3.2\%, and 3.4\% higher, respectively.
		
		% Please add the following required packages to your document preamble:
		% \usepackage{booktabs}
		\begin{table}[]
			\centering
			\fontsize{10}{15}\selectfont    %{字体尺寸}{行距}
			\caption{Comparison of HAFPN with other feature pyramid network when using the same backbone on solder joint defect dataset}
			
		    \resizebox{1.0\linewidth}{!}{
			\begin{tabular}{@{}lllllllllll@{}}
				\toprule
				
				& Backbone     & \multicolumn{3}{l}{Precision/\%}    & \multicolumn{3}{l}{Recall/\%}       & \multicolumn{3}{l}{mAP/\%}          \\ \cmidrule(lr){3-11}
				&              & Insufficient & Shifting & all  & Insufficient &  Shifting & all  & Insufficient & Shifting & all  \\ \cmidrule(lr){1-11}
				FPN                   & CSPDarknet53 & 91.4         & 82.5          & 87   & 98.5         & {\bf 82.6}          & 90.6 & 97.7         & 79.3          & 88.5 \\
				PAFPN                 & CSPDarknet53 & 87.7         & 75.0          & 81.4 & 98.6         & 73.9          & 86.3 & 96.5         & 77.9          & 87.2 \\
				ASFF                  & CSPDarknet53 & {\bf 91.6}         & 87.5          & 89.5 & 99           & 81.3          & 90.4 & {\bf 98.6}         & 82.6          & 90.6 \\
				BiFPN                 & CSPDarknet53 & 88.1         & 74.1          & 81.1 & 97.2         & 82            & 89.6 & 95.2         & 81.4          & 88.3 \\
				CFPNet                & CSPDarknet53 & 88.6         & 79.2          & 83.9 & 97.2         & {\bf 82.6}          & 89.9 & 93.1         & 83.1          & 88.1 \\
				{\bf HAFPN(ours) } & CSPDarknet53 & 91.4         & {\bf 90.2}          & {\bf 90.8} & {\bf 99.5}         & {\bf 82.6}          & {\bf 91.1} & 97.8         & {\bf 85.2}          & {\bf 91.5} \cr
				\bottomrule
				
			\end{tabular}
		}
		\end{table}

		This study used HAFPN to improve the YOLOv5 defect detection model and compares the improved defect detection model with different detection models on the solder joint defect dataset. The comparative models include one-stage detection models such as YOLOv4 \cite{29}, YOLOv5 \cite{30}, YOLOv7 \cite{31}, and YOLOv8 \cite{32}, improved YOLOv5 detection models such as STC-YOLOv5, TPH-YOLOv5, and two-stage detection models Faster R-CNN \cite{33}. Table 3 records the experimental results. Compared with the YOLO series algorithms, our model achieves the best overall precision, recall, and mAP indicators. Regarding detection speed, although FPS is lower than the original YOLOv5 model but higher than other models, its precision, recall, and mAP are 9.4\%, 4.8\%, and 4.3\% higher than YOLOv5. Compared to Faster R-CNN, the Recall value is lower, but the speed is three times faster, and the proposed algorithm has effective real-time performance. Compared with the improved YOLOv5 model STC-YOLOv5 and TPH-YOLOv5, the precision increased by 6.4\%, 2.4\%, recall increased by 3.1\%, 2.2\%, mAP increased by 2.8\%, 0.6\%, and FPS increased by 22.5, 31.6, respectively.
		
		\begin{table}[]
			\centering
			\fontsize{10}{15}\selectfont    %{字体尺寸}{行距}
			\caption{Comparative experiment on detection performance with different defect detection algorithms}
			
			\resizebox{1.0\linewidth}{!}{
				\begin{tabular}{@{}lllllllllll@{}}
					\toprule
					& \multicolumn{3}{l}{Precision/\%}              & \multicolumn{3}{l}{Recall/\%}                 & \multicolumn{3}{l}{mAP/\%}                    & FPS            \\ \cmidrule(lr){2-10}
					& Insufficient  & Shifting & all           & Insufficient  & Shifting & all           & Insufficient  & Shifting & all           &                \\
					\cmidrule(lr){1-11}
					YOLOv4                       & 82.3          & 70.1          & 76.2          & 87.4          & 59.8          & 73.6          & 90.6          & 72.2          & 81.4          & 123.5          \\
					YOLOv5                       & 87.7          & 75.0          & 81.4          & 98.6          & 73.9          & 86.3          & 96.5          & 77.9          & 87.2          & \textbf{163.9} \\
					YOLOv7                       & 87.2          & 56.2          & 71.7          & \textbf{99.5} & 78.3          & 89.1          & 95.6          & 67.9          & 81.7          & 140.5          \\
					YOLOv8                       & 87.8          & 68.5          & 78.2          & 98.2          & 60.9          & 79.7          & 96.6          & 71.7          & 84.2          & 113.6          \\
					Faster R-CNN                 & 87.3          & 82.0          & 84.7          & 98.7          & \textbf{91.5} & \textbf{95.1} & 90.1          & 75.3          & 82.7          & 50.6           \\
					STC-YOLOv5 {[}27{]}          & 90.8          & 77.9          & 84.4          & 99.2          & 76.7          & 88.0          & 96.5          & 80.9          & 88.7          & 137.3          \\
					TPH-YOLOv5 {[}26{]}          & 91.1          & 85.7          & 88.4          & \textbf{99.5} & 78.3          & 88.9          & 97.2          & 84.5          & 90.9          & 128.2          \\
					\textbf{YOLOv5+HAFPN (ours)} & \textbf{91.4} & \textbf{90.2} & \textbf{90.8} & \textbf{99.5} & 82.6          & 91.1          & \textbf{97.8} & \textbf{85.2} & \textbf{91.5} & 159.8 \cr
					\bottomrule        
				\end{tabular}
			}

		\end{table}
		
		We used the improved YOLOv5 network to visually compare the detection performance with the original YOLOv5 network, as shown in Figure 8. Out of 12 pins, the first nine are defective. It can be found that the original YOLOv5 network has missed detection for shifting defects (the first two pins) for smaller insufficient defect targets. The improved network detection capability has been enhanced, avoiding the occurrence of missed and false detections. In Figure 8(b), all defects have been correctly detected, and better detection results have been achieved. 
		
		\begin{figure}[!t]
			\centering
			\subfigure[]{
				\includegraphics[scale=0.55]{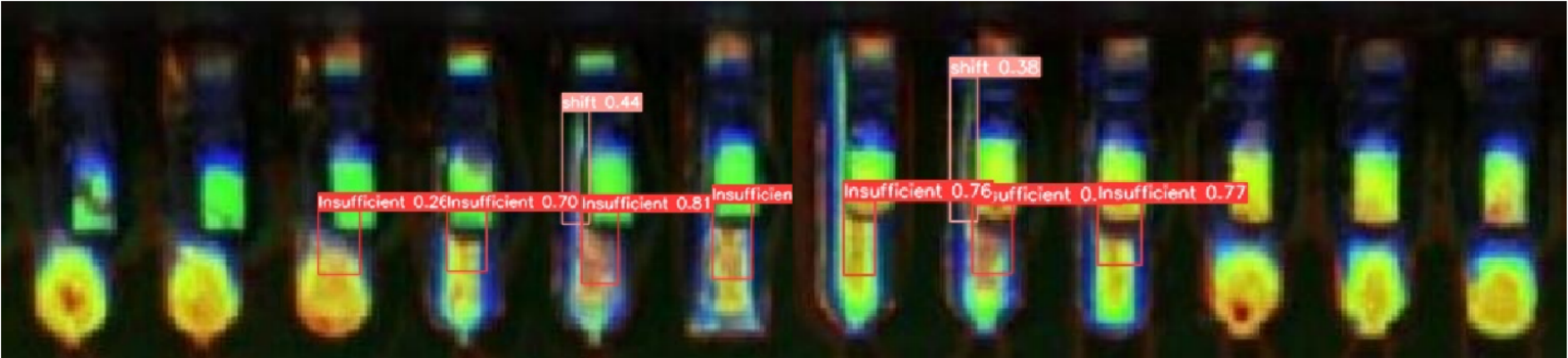}}\\
			\subfigure[]{
				\includegraphics[scale=0.55]{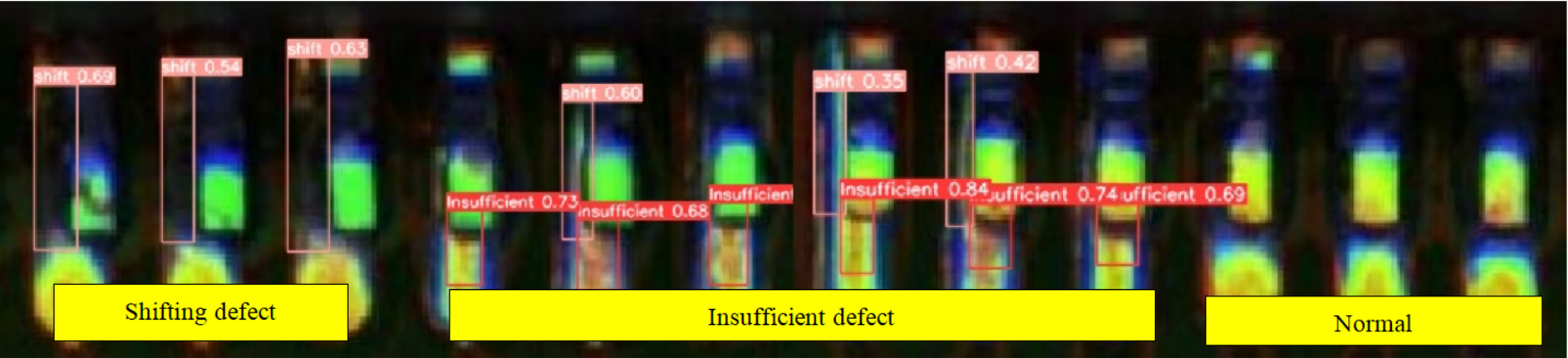}}
			\caption{Comparison of defect detection visualization effects between YOLOv5 (a) and YOLOv5+HAFPN (b)}
			\label{fig_8}
		\end{figure}

		(2) Ablation Study
		
		To verify the impact of attention module in the improved method on network detection performance, ablation experiments were designed using CSPDarknet53 as the backbone. The network detection effects of adding CA and EMSA in the two feature pyramids of FPN and PAFPN were compared, as shown in Table IV. When using FPN, adding EMSA to FPN resulted in an increase in precision, recall, and mAP by 3.3\%, 0.2\%, and 2.1\%, respectively. After adding the hybrid attention mechanism of EMSA and CA, precision, recall, and mAP increased by 3.5\%, 0.5\%, and 3\%, respectively. When using PAFPN, adding EMSA to PAFPN resulted in an increase in precision, recall, and mAP by 2.4\%, 4\%, and 1.7\%, respectively. After adding the hybrid attention mechanism of EMSA and CA, precision, recall, and mAP increased by 2.9\%, 4.2\%, and 3\%, respectively. The hybrid attention mechanism module has a certain improvement impact on the feature fusion performance of both feature pyramid networks, proving that the hybrid attention mechanism has stronger perception ability for the contextual information of features and can promote the improvement of defect detection performance.

		\begin{table}[]
			\centering
			\fontsize{8}{12}\selectfont    %{字体尺寸}{行距}
			\caption{Results of ablation experiments  on solder joint defect dataset}
			\begin{tabular}{@{}llllllll@{}}
				\toprule
				CSPDarknet53 & FPN & PAFPN & EMSA & CA & Precision/\% & Recall/\% & mAP/\% \\ \cmidrule(lr){1-8}
				\checkmark            & \checkmark   &       &      &    & 87           & 90.6      & 88.5   \\
				\checkmark            & \checkmark   &       & \checkmark    &    & 90.3         & 90.8      & 90.6   \\
				\checkmark            & \checkmark   &       & \checkmark    & \checkmark  & \textbf{90.8}         & \textbf{91.1}      & \textbf{91.5}   \\
				\checkmark            &     & \checkmark     &      &    & 81.4         & 86.3      & 87.2   \\
				\checkmark            &     & \checkmark     & \checkmark    &    & 83.8         & 90.3      & 88.9   \\
				\checkmark            &     & \checkmark     & \checkmark    & \checkmark  & 84.3         & 90.5      & 90.2 \cr
				\bottomrule 
			\end{tabular}
		\end{table}

		\section{CONCLUSIONS}
		In order to improve the defect detection accuracy of SMT solder joints in industrial scenarios and reduce the issues of missed detection and false alarm rates of defective solder joints, an enhanced multi-head self-attention mechanism is proposed by deep learning for defect detection method of SMT solder joint, which improve the network utilization range of features, and enable the network to have more robust nonlinear expression capabilities. We combine the CA mechanism with the EMSA mechanism to construct a hybrid attention mechanism network. The hybrid attention mechanism is employed to enhance FPN, improving its capability to fuse features, and enhancing information transmission between network channels. The enhanced FPN is applied to the YOLOv5 model, which improves YOLOv5's detection ability for solder joint defects, especially addressing the low detection rate issue for small-sized defects, while enhancing the generalization capability of the defect detection model. The method designed in this article enhances the feature fusion capability of the network through an improved attention mechanism. The experimental results show that our method achieves a mAP of 91.5\% on the solder joint defect dataset, which 4.3\% more expensive than the comparison model. Compared with the popular self attention improvement models STC-YOLO and TPH-YOLOv5, the mAP is 2.8\% and 0.6\% higher, and the FPS index is 159.8, which is 22.5 and 31.6 higher than STC-YOLO and TPH-YOLOv5, respectively. This FPS indicate shows that our model performs well in real-time and is beneficial for the application of industrial scenarios. The next step will be to continue improving the network to make it more lightweight in model parameters, while further improving the detection accuracy of solder joint defects.
		
		%\section*{Acknowledgment}
		%\noindent 
		
		%% the following bibliography is gererated manually for the sake of brevity
		%% only; please use a separate .bib file in your submission

		\bibliography{ref.bib}
		\bibliographystyle{alphaurl}
		
	%	\begin{thebibliography}{123}

	%	\end{thebibliography}

	\end{document}